%% file: main.tex
\documentclass[lettersize,journal]{IEEEtran}
\usepackage{amsmath,amsfonts}
\usepackage{algorithmic}
\usepackage{array}
\usepackage[caption=false,font=normalsize,labelfont=sf,textfont=sf]{subfig}
\usepackage{textcomp}
\usepackage{stfloats}
\usepackage{url}
\usepackage{verbatim}
\usepackage{graphicx}

\usepackage[skip=2pt]{caption} 
\usepackage{shortcuts}
\usepackage{tikz}
\usepackage{amsmath}
\usepackage{bm}
\usepackage[ruled,vlined,linesnumbered]{algorithm2e}
\usepackage{setspace}
\usepackage[most]{tcolorbox}
\usepackage[table]{xcolor} 
\usepackage{multirow,multicol,makecell}
\usepackage{threeparttable}
\usepackage{booktabs}
\usepackage{enumitem}
\usepackage[normalem]{ulem}
\usepackage{pifont}

\def\BibTeX{{\rm B\kern-.05em{\sc i\kern-.025em b}\kern-.08em
    T\kern-.1667em\lower.7ex\hbox{E}\kern-.125emX}}
\usepackage{balance}
\begin{document}
\title{Shadow Queries for Private Retrieval in Vector Databases}

\author{
Xinguo Feng,
Zhongkui Ma,
Zihan Wang,
Chuan Yan,
Guowei Yang,
Alsharif Abuadbba,
Guangdong Bai
\IEEEcompsocitemizethanks{
\IEEEcompsocthanksitem Xinguo Feng, Zhongkui Ma, Zihan Wang, and Guowei Yang are with The University of Queensland, Brisbane, Australia.
E-mail: \{s.feng, zhongkui.ma, zihan.wang, guowei.yang\}@uq.edu.au.
\IEEEcompsocthanksitem Alsharif Abuadbba is with CSIRO's Data 61, Sydney, Australia.
E-mail: \{Sharif.Abuadbba\}@data61.csiro.au.
\IEEEcompsocthanksitem Chuan Yan and Guangdong Bai are with City University of Hong Kong, Hong Kong.
E-mail: \{c.yan, g.bai\}@cityu.edu.hk.
}
}



\maketitle

\input{sections/0_abstract}


\input{sections/1_intro}
\input{sections/2_background}
\input{sections/3_approach-1}
\input{sections/4_experiments}

\input{sections/5_related_work}
\input{sections/6_conclusion}

\bibliographystyle{plain}
\bibliography{references}

\input{sections/Appendix}

\end{document}

%% file: sections/0_abstract.tex
\begin{abstract}
    Large language models (LLMs) are increasingly embedded into modern applications, such as writing assistants, coding assistants, AI agents, and search engines. However, they often struggle to provide accurate domain-specific knowledge without costly re-training or fine-tuning. Systems built on information retrieval (IR), such as Retrieval-Augmented Generation (RAG), address this by combining dense retrievers with embedding models and cloud-based vector databases that store pre-computed embeddings in the cloud to integrate external knowledge at inference time at scale. Despite the technical advancements, these embeddings are vulnerable to embedding inversion attacks (EIAs), 
    allowing adversaries to reconstruct the original text. While existing defenses attempt to mitigate this threat by modifying embeddings~(e.g., adding noise or scaling), they often provide insufficient protection, either largely sacrificing utility or being easily bypassed.

    In this work, we propose a novel \textbf{semantic-decomposition} and \textbf{embedding-decoupling} defense mechanism, \codename (\textbf{\underline{sha}}dow \textbf{\underline{q}}uery generation), to mitigate EIAs in cloud-based vector databases. Unlike prior works, \codename introduces a \textit{shadow query generation} pipeline that decouples the direct mapping between the stored embeddings and the original documents. \codename is built upon a key insight: \textit{EIAs fundamentally rely on the strong coupling between embeddings and their underlying textual content}. \codename consists of two steps: a \textit{generation} step and an \textit{indexing} step. 1) In the generation step, a generative language model is employed to synthesize relevant \textit{shadow queries} for each target document, which cover diverse aspects and semantic facets of the document content, achieving semantic decomposition. 2) In the indexing step, the generated shadow queries are encoded into embeddings, which then replace the original document embeddings in the vector database, achieving embedding decoupling. Evaluation across diverse IR datasets demonstrates the remarkable 
    effectiveness of \codename in protecting privacy (as low as 0.2104 recovery rate, up to 19.50\% more defended tokens than baseline defense) and preserving utility (up to 0.7967 in MAP@10, up to 5.53\% improvement compared to baseline defense). \codename demonstrates a defense paradigm shift from modifying resulting embeddings to enabling semantic decomposition and embedding decoupling with shadow query generation to counteract EIAs.
\end{abstract}

%% file: sections/1_intro.tex
\section{Introduction}
Large language models (LLMs)~\cite{achiam2023gpt, team2023gemini, touvron2023llama, team2024gemma} are increasingly embedded into modern applications, such as writing assistants~\cite{grammarly}, coding assistants~\cite{codex, claude-code}, AI agents~\cite{hermes_agent, open_claw}, and search engines~\cite{google-search, bing-search}.
However, the underlying LLMs often struggle in delivering accurate and up-to-date knowledge in specialized domains without costly re-training or fine-tuning, given their vast parameter spaces. 
To address this limitation, solutions built on information retrieval (IR)~\cite{singhal2001modern}, such as Retrieval-Augmented Generation (RAG), emerge as a powerful paradigm that integrates external knowledge sources into LLMs at inference time. By retrieving relevant information from domain-specific corpora, RAG enables LLMs to ground their responses in factual and domain-specific knowledge, significantly improving adaptability and reliability. The core of RAG is \textit{dense retrievers}~\cite{karpukhin-etal-2020-dense, ni2021large}, which leverage embedding models~\cite{bert, roberta} to encode both queries and documents into a shared vector space, allowing semantically similar texts to be effectively matched. To further support large-scale deployment, \textit{vector databases}~\cite{han2023comprehensive} have become standard infrastructure in RAG. By storing pre-computed embeddings of the documents, it enables efficient similarity search at retrieval time. In practice, commercial cloud-based solutions for vector databases, such as Pinecone~\cite{pinecone} and Milvus~\cite{milvus}, are often employed to host and serve the embeddings in the cloud, as they offer specialized optimizations, elastic scalability, and managed infrastructure. 

Despite these technical advancements, a critical vulnerability remains: \textit{embeddings themselves may leak information on the encoded data.}
Recent studies~\cite{song2020information, morris2023text, li2023sentence} show that embeddings can be exploited through \textit{embedding inversion attacks} (EIAs), which allow adversaries to reconstruct the original text from its embedding, posing a significant threat to data privacy in RAG.
Such attacks are rooted in the strong semantic coupling between the embedding and its underlying text, which induces an invertible mapping from the embedding to the original text that EIAs can effectively exploit.
An adversary who gains access to the stored embeddings (e.g., through data breaches or insider threats on the cloud-based data host) can recover private documents by leveraging the rich semantics encoded in the embeddings.
As RAG increasingly relies on vector databases,
the need to understand and mitigate the EIA risks grows even more urgent.

Several defense strategies are proposed for EIAs, including adding random noise~\cite{morris2023text}, or applying a secret scaling factor~\cite{zhuang2024understanding}, to the embeddings. However, these methods exhibit notable limitations. In the case of embedding noise, only minimal perturbations can be applied to the embeddings to preserve reasonable retrieval utility, struggling to achieve a utility-privacy balance. 
For secret scaling, it multiplies both query and document embeddings with a secret scalar known only to the user. Unfortunately, such scaling can be nullified by simple vector normalization on the embeddings, causing the defense to have no effect in practice.
These shortcomings reveal a fundamental tension between retrieval effectiveness preservation and privacy protection, highlighting the need for a robust and practical defense against EIAs. \textit{This urgent need motivates our exploration for a novel defense mechanism that safeguards document privacy while balancing retrieval performance.}

\paragraph{Our work}
We propose a novel \textbf{semantic-decomposition} and \textbf{embedding-decoupling} defense mechanism, \codename (\textbf{\underline{sha}}dow \textbf{\underline{q}}uery generation), to mitigate EIAs in vector databases.
Unlike previous methods that directly modify document embeddings to obscure encoded semantic information, \codename introduces a \textit{shadow query generation} pipeline that breaks the direct mapping between embeddings and underlying documents.
This design is motivated by a key insight:
\textit{EIAs fundamentally rely on the strong coupling between embeddings and their underlying texts.}
To disrupt this dependency, we leverage a generative language model to synthesize a set of \textit{shadow queries} that cover diverse semantic facets of the original documents.
These shadow queries are formulated from a user-query perspective, whose embeddings are then used to replace the document embeddings in the vector database.
This substitution is effective in two ways: 1) By breaking down the overall semantics of the document into diverse facets, this enables a \textit{semantic decomposition}. A successful retrieval is achieved if the user query is semantically close to one of the shadow queries.
2) By replacing the document embeddings with the shadow query embeddings that only capture semantic aspects of the document, it enables an \textit{embedding decoupling} between the stored embeddings and the original documents. This significantly reduces the risk of document reconstruction while maintaining effective retrieval, as the stored embeddings no longer encode the entire document but only a semantic characteristic of it. 

\codename operates in two stages: a \textit{generation} step and an \textit{indexing} step.
1) In the generation step, a generative language model is employed to synthesize relevant \textit{shadow queries} for each target document.
These shadow queries resemble plausible user queries that may retrieve the document, which cover diverse aspects and semantic facets of the document content.
This is accomplished through a crafted prompt that guides the language model to explore different query perspectives, and a K-Means clustering algorithm to ensure diverse outputs.
2) In the indexing step, the generated shadow queries are encoded into embeddings, which then replace the document embeddings in the vector database.
These embeddings are indexed into the vector database in randomized order, which can be mapped back to the corresponding documents during retrieval time. 
By combining the two steps, \codename ensures that the stored embeddings cover important semantic characteristics of the documents while disrupting their direct semantic links, thereby protecting document privacy against EIAs while balancing retrieval effectiveness.



We comprehensively evaluate \codename from two perspectives: its \textit{defense effectiveness} against EIAs and its \textit{retrieval utility} preservation.
Experiments are conducted on widely used IR benchmark datasets spanning diverse domains.
For defense evaluation, we assess the attack success rate using a state-of-the-art EIA method~\cite{morris2023text}, considering both standard and adaptive adversarial scenarios.
\codename consistently outperforms existing countermeasures, resulting in a recovery rate of 0.2710 on average and as low as 0.2104 measured in ROUGE-1, reducing the number of recovered tokens by 12.43\% on average and up to 19.50\% compared to prior defenses.
Furthermore, in a challenging adaptive attack setting where the attacker has full knowledge of the defense mechanism and hyperparameters, \codename remains highly effective with a recovery rate of 0.3376 on average and as low as 0.2083 measured in ROUGE-1, demonstrating strong resilience to adaptive attack strategies.
For utility evaluation, \codename achieves a remarkable balance between privacy and utility among all compared methods.
It maintains retrieval performance on par with an undefended system, achieving an NDCG up to 0.8524. This confirms \codename's capability in defending against EIAs while balancing retrieval performance. Ablation studies on variable-length documents and a different embedding model show that \codename is a generic and model-agnostic defense, which can be directly integrated into modern dense retrievers.



\paragraph{Contributions}
Our contributions are summarized as follows:
\begin{itemize}[left=0pt]
    \item
    \textbf{A novel defense mechanism for vector databases privacy via semantic decomposition and embedding decoupling.}
    We propose \codename, a novel framework that mitigates EIAs in vector databases. Instead of modifying document embeddings, \codename replaces them with embeddings of \textit{shadow queries} that cover diverse aspects and semantic facets of the documents.
    This shifts the defense paradigm from direct embedding perturbation to decoupling the inherent links between embeddings and documents.
    \item
    \textbf{A generic and model-agnostic approach.}
    \codename adopts a two-stage pipeline, shadow query generation and indexing, which is easy to adopt and compatible with existing dense retriever architectures. It works seamlessly with variable-length documents and various embedding models, facilitating a one-time offline process that occurs before retrieval time. 
    \item
    \textbf{An extensive empirical validation.}
    We evaluate \codename on widely used IR benchmark datasets spanning diverse domains.
    Experiments show that \codename outperforms existing countermeasures on defense efficacy, while maintaining retrieval performance on par with undefended systems, achieving a strong privacy-utility balance. \codename remains effective and robust in challenging adaptive scenarios, where the adversary has significantly enhanced capabilities and knowledge. 
\end{itemize}

\paragraph{Availability} Our code is publicly available at: \url{https://github.com/shanefeng123/SHAQ}. 

%% file: sections/2_background.tex
\section{Background}
\label{sec:background}

In this section, we introduce the background knowledge necessary to facilitate a thorough understanding of \codename.
\subsection{Information Retrieval: From Sparse to Dense}
\label{bg: ir}

\paragraph{Sparse retrievers} Information retrieval systems~\cite{singhal2001modern} in Natural Language Processing (NLP) fetch relevant documents given a user query.
Traditionally, they rely on sparse representations, such as bag-of-words~\cite{zhang2010understanding} or term frequency-inverse document frequency (TF-IDF)~\cite{ramos2003using}, to measure surface-level lexical overlaps, and apply ranking methods, such as BM25~\cite{robertson2009probabilistic}.
These methods, although computationally efficient, fall short when queries and documents express similar semantics but differ in wording. 

\paragraph{Embedding models}
To effectively measure the semantics similarity between texts, efforts have been made to integrate embedding models to produce vector representations in the retrieval pipeline. 
Modern embedding models are based on the transformer architecture~\cite{transformer}. Specifically, the masked language modeling (MLM) paradigm (e.g., BERT~\cite{bert}, RoBERTa~\cite{roberta}) is largely employed.
These pre-trained models tokenize input text using methods such as WordPiece~\cite{song2020fast} or Byte Pair Encoding~\cite{shibata1999byte}, and transform the token sequence into a high-dimensional embedding matrix.
A vector representation is then derived via average pooling of the matrix or the selection of a special token's embedding (e.g., using the \texttt{[CLS]} token in BERT).

\paragraph{Dense retrievers} By employing embedding models, dense retrievers, such as DPR~\cite{karpukhin-etal-2020-dense} and GTR~\cite{ni2021large}, have emerged to address the limitations of sparse retrievers.
They embed both queries and documents into a continuous vector space, where semantically similar texts are close to each other.
Such dense representations capture semantics and contextual relationships, enabling more effective similarity comparison between queries and documents by vector operations.
An embedding model, denoted as $\phi(\cdot)$, encodes a piece of text (query or document) to a vector in a high-dimensional space.
Given a query $\bm{q}$ and a set of documents $\mathcal{D}$, retrieving the top-$k$ relevant documents $\mathcal{D}_{\mathcal{K}}$ ($\mathcal{K}$ denotes the set of indices of the top-$k$ documents and $|\mathcal{K}| = k$) can be formally described as 
$\mathcal{D}_{\mathcal{K}} = \arg \mathrm{topk}_{\bm{d} \in \mathcal{D}} \; \text{sim}(\phi(\bm{q}), \phi(\bm{d}))$,
where $\bm{d} \in \mathcal{D}$ denotes a document in the set of documents $\mathcal{D}$, and $\text{sim}(\cdot, \cdot)$ is a similarity function (e.g., cosine similarity or dot product) that measures the similarity between the query embedding $\phi(\bm{q})$ and the document embedding $\phi(\bm{d})$.

\paragraph{Vector databases}
Dense retrievers largely adopt vector databases to store pre-computed document embeddings, thereby avoiding costly re-encoding at retrieval time.
In practice, highly optimized cloud-based commercial solutions~\cite{pinecone, milvus} are typically employed.

\paragraph{Retrieval-Augmented Generation}
RAG~\cite{lewis2020retrieval} is a recently emerging technique that employs dense retrievers to provide LLMs with domain-specific knowledge before generation.
When the user prompts the LLM, the user's prompt is used as a query to retrieve relevant documents. The retrieved documents are then concatenated with the user prompt to be fed into the LLM for better-quality generation.

\subsection{Embedding Inversion Attacks and Defenses}
\label{bg: eia}

\paragraph{Embedding inversion attacks}
Although embeddings abstract away the surface form of text, they can still leak private information on the encoded data.
Embedding inversion attacks~\cite{song2020information, morris2023text, li2023sentence} are techniques designed to reconstruct the original text from its embedding by exploiting the inherent connections between them. These attacks assume that the adversary has access to the embeddings and the embedding model adopted.
Earlier attack~\cite{song2020information} employs a gradient-based optimization, where a dummy embedding is optimized to match the observed embedding.
Recently, more advanced attacks such as vec2text~\cite{morris2023text} and GEIA~\cite{li2023sentence}, train an attack generative language model to predict the original text conditioned on the observed embedding.
Specifically, the vec2text attack, focusing on document embeddings in vector databases, is the main risk this work aims to mitigate.


\paragraph{Defenses against embedding inversion attacks}
To mitigate the risks of EIAs, various defense mechanisms have been proposed.
A common strategy is via noise injection~\cite{morris2023text}, aiming to distort the semantics encoded in the embeddings to obscure the inversion process. However, only minimal perturbations can be applied to preserve an effective retrieval, struggling to achieve a utility-privacy balance.
Another line of work multiplies both the query and document embeddings by a secret scaling factor~\cite{zhuang2024understanding} known only to the user, aiming to disrupt the inversion process by scaling the embedding vectors. However, this can be easily bypassed by a simple vector normalization operation, making it has no effect in practice.
More importantly, these existing defenses directly modify the document embeddings, which limits their effectiveness due to the inherent semantic coupling between the embeddings and the underlying texts.

%% file: sections/3_approach-1.tex
\section{Approach}
\label{sec:approach}

This section presents a detailed description of \codename.
We begin by outlining the threat model in Section~\ref{sec:threat_model}, followed by a high-level overview of our defense mechanism in Section~\ref{sec:overview}.
The core components of \codename are described in Sections~\ref{sec:approach:generation} and~\ref{sec:approach:indexing}, respectively.

\subsection{Threat Model}
\label{sec:threat_model}

We consider a dense retrieval setting, where document embeddings are pre-computed and stored in a cloud-based vector database hosted by a third-party provider for efficient retrieval.
We assume an \emph{honest-but-curious} adversary who has access to the stored embeddings, either through insider threats or data breaches in the data host. The adversary can also access the employed embedding model in a black-box manner, meaning that the adversary can obtain the embeddings of any arbitrary texts.
The adversary's objective is to reconstruct the original documents associated with the observed document embeddings in the vector database.
This threat model aligns with those used in the prior EIA study~\cite{morris2023text}, and reflects practical deployment concerns with a cloud-based commercial vector database solution.

\subsection{Overview}
\label{sec:overview}

\input{figures/overview}

Figure~\ref{fig:overview} illustrates the overall architecture of \codename, which comprises two core components:

\begin{enumerate}[left=0pt]
    \item \textbf{Shadow query generation}: A generative language model is employed to synthesize relevant shadow queries for each target document.
    These shadow queries resemble plausible user queries that may retrieve the document, which cover diverse aspects and semantic facets of the document content, achieving semantic decomposition.
    This is accomplished through a crafted prompt that guides the language model to explore different query perspectives, and a K-Means clustering algorithm to ensure diverse outputs.
    \item \textbf{Embedding indexing}: The generated shadow queries are encoded into embeddings, which then replace the document embeddings in the vector database, achieving embedding decoupling.
    These embeddings are indexed into the vector database in randomized order, accompanied by a mapping from shadow queries to their corresponding documents.
\end{enumerate}

By combining the two steps, \codename ensures that the stored embeddings cover important semantic characteristics of the original document while removing their direct semantic links, thereby protecting document privacy against EIAs while balancing retrieval effectiveness. We detail each step as follows.

\subsection{Shadow Query Generation}
\label{sec:approach:generation}

\input{algos/algo1-1}

The generation step in \codename synthesizes shadow queries that cover different aspects and semantic facets of a target document, resembling potential user queries that may retrieve it. 
This draws inspiration from doc2query~\cite{nogueira2019document}, where a generated query is appended to the document to improve the retrieval performance.
By decomposing the semantics of the document into several shadow queries, direct exposure of the document's content is avoided.
This is accomplished by employing a generative language model for shadow queries generation and the K-Means clustering algorithm for semantics diversity.
The resulting shadow queries are then encoded into embeddings, which replace the document embeddings during the indexing step.

Algorithm~\ref{alg:generation} outlines the generation process (\texttt{Generate}).
It iterates over each document $\bm{d}^{(i)}$ in the collection $\mathcal{D}$ (lines 1--2).
Given a document $\bm{d}^{(i)}$ and the desired number of query candidates $n_g$, it first constructs a prompt $\bm{p}$ using a carefully designed prompt template (\texttt{ConstructPrompt} in line 3).
This template is crafted to guide the language model in producing a diverse and concise set of shadow queries that capture various semantic facets of the input document.
The constructed prompt $\bm{p}$ is then fed into the language model $f$ to produce a set of $n_g$ candidate shadow queries $\mathcal{Q}^{\text{gen}}$ (line 4).
To encourage comprehensive coverage of diverse semantic aspects, it clusters the generated queries into $n_k$ groups using K-Means clustering (\texttt{KMeans} in line 5).
From each cluster $\mathcal{C}_j$, one shadow query is randomly sampled (\texttt{RandomChoice}) to form the shadow query set $\mathcal{Q}^{\text{sel}}$ (lines 6--9).
These selected queries are aggregated into the final set $\mathcal{Q}$ (line 10) and passed to the subsequent indexing step.

The prompt template used in this step and an example of the generated shadow queries are presented in Appendix~\ref{prompt_template} and ~\ref{generated_shadow_queries}.

\subsection{Embedding Indexing}
\label{sec:approach:indexing}

\input{algos/algo2-1}

The goal of the indexing step in \codename is to construct a vector database that replaces the document embeddings with those of the shadow queries. By replacing the embeddings, the directly reconstructed text from an observed embedding by EIAs is no longer the original document. At the same time, as these embeddings still encode a semantic facet of the original documents, an effective retrieval can be facilitated.

The indexing procedure is detailed in Algorithm~\ref{alg:indexing}.
For each document $\bm{d}$, the algorithm first selects its associated shadow queries $\mathcal{Q}^d$ (line 3--5).
Then, for each shadow query $\bm{q}_j$, it gets encoded into an embedding $\bm{e}$ with the embedding model $\phi$ (line 7), and the corresponding document ID is recorded in a mapping $\mathcal{S}$ for retrieval purposes (line 8).
The embedding is then added to the vector database $\mathcal{V}$ (line 9).
After all shadow queries have been processed, the embeddings in the vector database are randomly shuffled (line 10).
Finally, the algorithm returns the embedding-to-document-ID mapping $\mathcal{S}$ and the vector database $\mathcal{V}$.

\paragraph{Scattering indexing strategy}
We index the embeddings in a vector database for efficient retrieval.
Recall from the generation step that each document is associated with $n_k$ shadow queries.
Instead of aggregating the embeddings of the shadow queries into a single pooled embedding, which we find to degrade retrieval utility due to semantic interference in various shadow queries, \codename employs a scattering strategy.
In particular, the order of the shadow query embeddings corresponding to each document is randomly shuffled in the vector database (line 10).
This random scattering strategy, rather than sequential indexing, is designed to mitigate potential adaptive attacks, where the attacker may easily identify all the shadow query embeddings corresponding to the same document. This is elaborated later in Section~\ref{adaptive}.

%% file: figures/overview.tex
\begin{figure}[!t]
    \centering
    \includegraphics[width=0.85\linewidth]{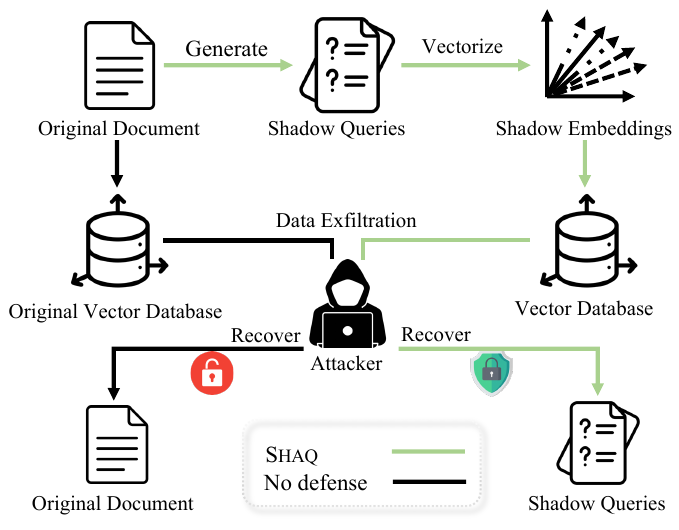}
    \caption{Overview of \codename.}
    \label{fig:overview}
\end{figure}

%% file: algos/algo1-1.tex
\begin{algorithm}[!htbp]

\caption{\texttt{Generate($\cdot$)}}
\label{alg:generation}

\setstretch{0.8}

\KwIn{
    $f$ -- the generative language model;
    $\mathcal{D} = \{\bm{d}^{(1)}, \ldots, \bm{d}^{(N)}\}$ -- the collection of documents;
    $n_g$ -- number of shadow query candidates per document;
    $n_k$ -- number of selected shadow queries per document.
}
\KwOut{
    $\mathcal{Q}$ -- the list of generated shadow queries.
}
$\mathcal{Q} \gets \texttt{list()}$ \;

\ForEach{$\bm{d}^{(i)} \in \mathcal{D}$}{
    $\bm{p}$ $\gets$ \texttt{ConstructPrompt($\bm{d}$, $n_g$)} \;
    
    $\mathcal{Q}^{\text{gen}}$ $\gets$ $f(\bm{p})$ \;
    
    $\mathcal{C}$ $\gets$ \texttt{KMeans($\mathcal{Q}^{\text{gen}}$, $n_k$)} \;
    
    $\mathcal{Q}^{\text{sel}}$ $\gets$ \texttt{list()} \;
    
    \ForEach{$\mathcal{C}_j$ $\in$ $\mathcal{C}$}{
        $\bm{q}$ $\gets$ \texttt{RandomChoice($\mathcal{C}_j$)} \;
        
        \texttt{$\mathcal{Q}^{\text{sel}}$.append($\bm{q}$)} \;
    }
    \texttt{$\mathcal{Q}$.extend($\mathcal{Q}^{\text{sel}}$)} \;
}

\Return $\mathcal{Q}$
\end{algorithm}

%% file: algos/algo2-1.tex
\begin{algorithm}[!htbp]

\caption{\texttt{Index}$(\cdot)$}
\label{alg:indexing}

\setstretch{0.8}

\KwIn{
    $\phi$ -- the embedding model;
    $\mathcal{Q}$ -- the list of all shadow queries;
    $\mathcal{D} = \{\bm{d}^{(1)}, \ldots, \bm{d}^{(N)}\}$ -- the document collection;
    $n_k$ -- number of shadow queries selected per document.
}
\KwOut{
    $\mathcal{S}$ -- mappings between shadow query embeddings and document IDs;
    $\mathcal{V}$ -- the resulting vector database of shadow query embeddings.
}

$\mathcal{S}$ $\gets$ \texttt{dict()} \;

$\mathcal{V}$ $\gets$ \texttt{list()} \;

\For{\texttt{$i$ $\in$ range($0$, $\mathcal{D}$.length)}}{
$\bm{d} = \mathcal{D}[i]$ \;
    
    $\mathcal{Q}^d$ $\gets$ $\mathcal{Q}$[$i \cdot n_k$ : $(i + 1) \cdot n_k$] \;
    
    \ForEach{$\bm{q}_j$ $\in$ $\mathcal{Q}^d$}{
        $\bm{e}$ $\gets$ $\phi(\bm{q}_j)$ \;
        
        \texttt{$\mathcal{S}$[$\bm{e}$] $\gets$ id($\bm{d}$)} \;
        
        \texttt{$\mathcal{V}$.append($\bm{e}$)} \;
    }
}

\texttt{$\mathcal{V}$ $\gets$ RandomShuffle($\mathcal{V}$)} \;

\Return $\mathcal{S}, \mathcal{V}$
\end{algorithm}


%% file: sections/4_experiments.tex
\section{Experimental Evaluation}
This section details the evaluation of \codename. Section~\ref{exp_settings} introduces the experiment settings.
Section~\ref{proof-of-concept} provides a proof-of-concept experiment to demonstrate the semantic decomposition and embedding decoupling ability of \codename. 
Section~\ref{utility} and ~\ref{defense} present the retrieval utility and defense efficacy under various defenses against EIA. Section~\ref{ablation} presents ablation studies with variable-length documents and a different embedding model.
Section~\ref{adaptive} demonstrates \codename's robustness against a challenging adaptive attack scenario.

\subsection{Experimental Settings}
\label{exp_settings}

This section introduces the experimental settings.
Section~\ref{exp_settings:baselines} introduces the baselines of attack and defenses. Section~\ref{exp_settings:ir_systems} presents the settings of the dense retriever. Section~\ref{exp_settings: metrics} introduces the metrics used in the evaluation. Section~\ref{exp_settings: our_method} lists the settings of \codename. 
Additional discussion on the empirical implementation is presented in Appendix~\ref{appendix: implementation}.

\subsubsection{Baselines of Attack and Defenses}
\label{exp_settings:baselines}
We introduce the baselines of attack and defenses below.

\paragraph{Baseline attack}
For EIA, we adopt the state-of-the-art vec2text attack~\cite{morris2023text} as our attack baseline, which specifically targets document embeddings in vector databases in modern dense retrievers, and achieves substantial attack efficacy. The vec2text attack employs generative language models based on the T5 model~\cite{raffel2020exploring} as their attack models. These models are trained to reconstruct texts from embeddings, which take an embedding as input and generate the corresponding text. They are trained with embedding-text pairs derived from the training split of either the Natural Questions dataset~\cite{kwiatkowski2019natural} or the MS MARCO dataset~\cite{nguyen2016ms}, where the documents are truncated to 32 tokens for simplicity.
Throughout our experiments, we focus on the attack model trained with embeddings produced by the open-source embedding model, GTR-T5 encoder~\cite{ni2021large}. Although vec2text also includes an attack model trained on embeddings produced by OpenAI's text-embeddings-ada-002 model~\cite{openai2022textembeddingada002}, without loss of generality, we exclude this setting due to the commercial model's usage cost, given the large scale of our evaluation. We also exclude the previous attacks~\cite{song2020information, li2023sentence} from our evaluation, as they are not for EIAs in vector databases, or not for document-level recovery.

\paragraph{Baseline defenses}
For defenses, we consider embedding noise~\cite{morris2023text} and the secret scaling factor~\cite{zhuang2024understanding} as baselines, 
which are proposed and adopted to counter EIAs in previous studies.
However, the secret scaling defense is easily circumvented by a simple vector normalization, effectively nullifying its effect in practice. As such, we consider this defense as equivalent to having no defense applied.

\subsubsection{Settings of Dense Retriever}
\label{exp_settings:ir_systems}
We introduce the settings of dense retriever below.

\paragraph{Task} We focus on the standard information retrieval task that aims to retrieve relevant documents given a user query using dense retrievers, where a vector database is employed to store the pre-computed document embeddings. For each query, documents in the corpus are ranked based on the similarities (cosine similarity or dot product) between the stored embeddings and the query embedding. The top-$k$ similar documents are retrieved as relevant documents, either via direct retrieval or with embedding-to-document-ID mapping if \codename is applied. We employ the FAISS~\cite{douze2024faiss} library for efficient similarity comparison. The effectiveness of the downstream tasks, such as question answering or text generation based on the retrieved documents, is beyond the scope of this evaluation.

\paragraph{Embedding model} 
To ensure the embedding compatibility with the attack model, we adopt the same GTR-T5 encoder as our embedding model throughout the experiments, which outputs normalized embeddings and uses cosine similarity as the similarity objective. 

\input{tables/datasets}

\paragraph{Datasets} We evaluate \codename on a range of IR datasets from the widely used BEIR benchmark~\cite{thakur2021beir}, which cover diverse domains such as scientific, financial, and open-domain question answering. Each dataset includes queries, documents, and ground truth relevance labels that indicate the relevant documents for each query. The development or test splits of these datasets are used in the evaluation to avoid potential information leakage in the original training of the attack model.

The BEIR benchmark contains datasets with varying characteristics, including the number of queries (49 to 6,980), documents (5,183 to 8,841,823), average relevant documents per query (from 1 to 493.5), and relevance labels (binary to multi-level). 
To ensure a consistent evaluation process, we exclude datasets with non-binary relevance labels or an excessively large number of relevant documents per query.
Considering the large scale of the evaluation, we randomly sample 100 queries for each of the evaluated datasets.
For each sampled query, we include all relevant documents, then augment the document corpus with the top-100 ranked documents using BM25~\cite{robertson2009probabilistic} to include irrelevant documents, following the subsampling strategy introduced by Fr{\"o}be et al.~\cite{frobe2025corpus}. 
Note that since a query may have multiple relevant documents, and the same document may be relevant to multiple queries, the number of sampled documents varies across datasets.

\input{tables/data_obscuration}

The statistical details of the evaluated datasets, e.g., the numbers of original queries and documents, the numbers of sampled queries and documents, and the original average documents per query, are presented in Table~\ref{table:datasets}. More comprehensive details of the evaluated datasets can be found in Appendix~\ref{appendix: dataset details}. 
The datasets removed from our evaluation are summarized in Table~\ref{table:removed_datasets} in Appendix~\ref{appendix: removed_datasets}.

\paragraph{Data preprocessing} Since the attack model in vec2text is trained to predict texts with a fixed length of 32 tokens, we truncate all sampled documents to 32 tokens before the shadow query generation to match this setup and to avoid giving the defender an unfair advantage, as longer texts are harder to recover.

\subsubsection{Evaluation Metrics} We introduce the evaluation metrics below.
\label{exp_settings: metrics}

\paragraph{Metrics on retrieval utility} We employ several widely used IR metrics to assess retrieval effectiveness, including NDCG~\cite{jarvelin2002cumulated}, MAP~\cite{voorhees1999trec}, Recall, Precision, and Accuracy. Following standard practice, we adopt a top-$k$ evaluation with $k=10$, aligning with the evaluation in vec2text~\cite{morris2023text}. All utility metrics range from 0 to 1, with higher values indicating better retrieval performance.

\paragraph{Metrics on defense efficacy} We employ various widely used lexical-level metrics to evaluate defense efficacy, including BLEU~\cite{papineni2002bleu}, ROUGE~\cite{lin2004rouge}, and METEOR~\cite{banerjee2005meteor}. All metrics range from 0 to 1, with lower values indicating less similarity between the recovered documents and the original documents, hence better defense efficacy.

\subsubsection{Settings of \codename} We introduce the settings of \codename below.
\label{exp_settings: our_method}

\paragraph{Generative language model}
For shadow query generation, we utilize the QwQ-32B language model~\cite{qwq32b}, which has demonstrated strong performance across a wide range of NLP tasks. Its high generation quality and capacity make it well-suited for producing diverse shadow queries. We use this model with an 8-bit quantization to loosen the memory requirement.

\paragraph{Hyperparameters} We conduct a hyperparameter search with the Scifact dataset for a key hyperparameter: $n_k$, the number of final shadow queries per document in \codename. We select $n_k=10$ as it yields the best utility and privacy balance. To ensure a fair comparison with baseline defenses, we also run a hyperparameter search for $\sigma$ in embedding noise, and select $\sigma = 0.01$, as it yields comparable utility with $n_k=10$ in \codename. 
The details of the hyperparameter search are presented in Appendix~\ref{hyper}. By systematically tuning these parameters to achieve comparable retrieval utility, we ensure a fair comparison of the defense efficacy between the two defenses throughout the experiments.

\input{tables/utility}
\input{tables/privacy}

\subsection{Demonstrating Semantic Decomposition and Embedding Decoupling}
\label{proof-of-concept}
We begin with a proof-of-concept experiment to demonstrate how \codename achieves semantic decomposition and embedding decoupling.

\subsubsection{Semantic Decomposition}
To assess semantic decomposition, we consider both embedding-level and lexical-level similarities. At the embedding level, we concatenate all shadow queries corresponding to a document and compute the resulting embedding of the concatenated text. 
We then compute the cosine similarity between the resulting embedding and the document embedding. At the lexical level, we measure the proportion of document tokens that appear across its shadow queries. These evaluations capture how well the collection of shadow queries preserves the overall semantics of the original document.

\paragraph{Results} Table~\ref{table:obscuration} shows the results. At the embedding level, the concatenated shadow queries achieve high similarity with the original document embeddings (up to 0.9162), indicating that the overall semantics of the document are well preserved. At the lexical level, token coverage is also high, reaching a value of 0.7980, indicating that most document tokens are represented across the shadow queries. These results confirm that the collection of shadow queries effectively captures most of the semantic content of the document, validating the semantic decomposition property of \codename.

\subsubsection{Embedding Decoupling}
To evaluate embedding decoupling, we again examine both embedding-level and lexical-level similarities between documents and shadow queries. At the embedding level, for each target document, we compute the mean cosine similarity between its original embedding and the embeddings of the corresponding shadow queries. At the lexical level, we measure the textual similarity between the document and its shadow queries using the defense efficacy metrics introduced in Section~\ref{exp_settings: metrics}. These comparisons assess whether individual shadow queries remain semantically related to the document while preventing direct reconstruction of its full content.

\paragraph{Results} Table~\ref{table:obscuration} summarizes the findings. At the embedding level, shadow query embeddings show relatively high similarity with their corresponding document embeddings (up to 0.8310), indicating that they capture meaningful semantic facets for effective retrieval. In contrast, at the lexical level, the overlaps are minimal, with similarity scores as low as 0.0023 (BLEU), 0.1949 (R-1), 0.0489 (R-2), 0.1561 (R-L), and 0.1015 (METEOR). These results confirm that while shadow queries preserve sufficient semantic signals for effective retrieval, recovering an individual shadow query does not expose the full content of the original document. These results validate the embedding decoupling property of \codename.

\subsection{Retrieval Utility Preservation}
\label{utility}
We assess the retrieval utility preservation of \codename by analyzing its retrieval performance. Specifically, for each query, it is encoded into an embedding vector, and the cosine similarities with all shadow query embeddings in the vector database are computed. The shadow queries in the retrieval rank (top-10 similar) are mapped back to the original documents using the embedding-to-document-ID mapping, and the documents are retrieved accordingly. If the retrieved documents are the relevant documents, it represents a successful retrieval. This evaluation is conducted across multiple datasets and compared against different baseline defenses, where the baseline defenses follow a standard embedding similarity comparison between the query embeddings and the original document embeddings.

\paragraph{Results} Table~\ref{table:utility} presents the retrieval utility results. Overall, both defense strategies deliver comparable performance with the undefended system across various datasets and metrics. When no defense is applied, the retrieval achieves NDCG from 0.1876 to 0.9146, MAP from 0.0705 to 0.8745, Recall from 0.1107 to 0.9822, Precision from 0.0370 to 0.1550, and Accuracy from 0.3500 to 1.0000. 
When embedding noise or \codename is applied, only minimal performance variations are observed.
For embedding noise, the variations from the undefended performance range from $-0.0131$ to $+0.0076$ for NDCG, $-0.0163$ to $+0.0091$ for MAP, $-0.0200$ to $+0.0350$ for Precision, $-0.0040$ to $+0.0040$ for Recall, and $-0.0200$ to $+0.0400$ for Accuracy. 
For \codename, the variations from the undefended performance range from $-0.0747$ to $+0.0382$ for NDCG, $-0.0778$ to $+0.0390$ for MAP, $-0.1100$ to $+0.0640$ for Recall, $-0.0110$ to $+0.0110$ for Precision, and $-0.1100$ to $+0.0600$ for Accuracy.

These results demonstrate that \codename maintains competitive retrieval effectiveness across diverse domains, on par with an undefended system or when embedding noise is applied, showcasing its ability to preserve retrieval utility. 

\subsection{Defense Efficacy}
\label{defense}
We evaluate the defense efficacy of \codename against vec2text, the state-of-the-art EIA in vector databases. Specifically, the reconstructed text from an embedding is compared to its corresponding original document. We sample 100 indexed embeddings from the vector database for evaluation. In particular, for the undefended and embedding noise baselines, we sample 100 document embeddings as attack targets. For \codename, we sample 100 shadow query embeddings, where each embedding corresponds to a shadow query generated from a distinct document. This setup reflects a realistic scenario in which the attacker attempts to recover the original document from a single embedding, given that the attacker is not aware of the defense mechanism of \codename. A challenging adaptive attack scenario, where the adversary has complete knowledge of the defense mechanism, and attempts to recover a document using multiple shadow query embeddings, is discussed in Section~\ref{adaptive}.

\paragraph{Results} We present the results in Table~\ref{table:privacy}. In the undefended setting, the reconstructed texts achieve scores ranging from $0.3052$ to $0.9302$ (BLEU), $0.6854$ to $0.9794$ (R-1), $0.5013$ to $0.9429$ (R-2), $0.6282$ to $0.9646$ (R-L), and $0.6635$ to $0.9765$ (METEOR), indicating that a substantial portion of the original documents can be accurately recovered. This underscores the severity of the privacy risk posed by EIAs. 
With the embedding noise defense applied, reconstruction accuracy is significantly reduced compared to the undefended system, with variations from $-0.2653$ to $-0.8581$ (BLEU), $-0.3127$ to $-0.5842$ (R-1), $-0.3728$ to $-0.8182$ (R-2), $-0.3541$ to $-0.6821$ (R-L), and $-0.3036$ to $-0.6828$ (METEOR). While the attack efficacy is notably decreased, a relatively large portion of document content can still be recovered in most cases, leaving residual privacy concerns. 
\codename achieves the lowest reconstruction scores across all metrics and datasets. Comparing to the undefended system, the variations ranging from $-0.2686$ to $-0.9059$ (BLEU), $-0.4176$ to $-0.7213$ (R-1), $-0.4032$ to $-0.8207$ (R-2), $-0.3868$ to $-0.7235$ (R-L), and $-0.4459$ to $-0.8168$ (METEOR). It consistently outperforms the defense efficacy of embedding noise by up to $-0.0548$ in BLEU, $-0.1949$ in R-1, $-0.0718$ in R-2, $-0.1004$ in R-L, and $-0.1689$ in METEOR, demonstrating strong privacy protection against EIAs.

These results show that \codename offers a strong privacy-utility trade-off. While maintaining retrieval performance comparable to existing baselines, it significantly reduces the privacy risk of EIAs, making it a practical and effective defense for embeddings in vector databases.

\subsection{Ablation Studies}
\label{ablation}

We conduct ablations to isolate the factors that may affect \codename's performance. Specifically, we evaluate retrieval utility (i) when the documents are untruncated and (ii) when the embedding model uses an alternative similarity measure. In both settings, \codename maintains robust retrieval utility, demonstrating its generality on variable-length documents and model-agnostic ability. Full details of the ablation studies are presented in Appendix~\ref{appendix:ablation}.

\subsection{Adaptive Attack}
\label{adaptive}
Beyond the standard EIA, we evaluate the robustness of \codename against a challenging adaptive attack, where the attacker has significantly enhanced knowledge of the defense mechanism, its parameters, and has access to the LLM employed for shadow query generation. More importantly, the attacker knows that the observed embeddings are those of the shadow queries for the corresponding target documents, and designs a sophisticated adaptive attack strategy. In particular, the attacker aims to first recover all shadow queries for the target document, then attempts to piece together the original document. \codename demonstrates remarkable robustness in this challenging adaptive attack scenario, limiting the attack efficacy to as low as 0.2083 in R-1. The details of the adaptive attack are presented in Appendix~\ref{appendix: adaptive}.

%% file: tables/datasets.tex
\begin{table}[!htbp]

\centering

\caption{Datasets split and statistical details.}
\label{table:datasets}

\renewcommand{\arraystretch}{.8}
\setlength{\tabcolsep}{0.5pt}

\begin{threeparttable}
\resizebox{\linewidth}{!}{
\begin{tabular}{
l@{\hspace{0pt}}
rrrrr
}

\toprule
\makecell[c]{Dataset} &
\makecell[c]{\#Original\\Queries} &
\makecell[c]{\#Original\\Documents} &
\makecell[c]{\#Sampled\\Queries} &
\makecell[c]{\#Sampled\\Documents} &
\makecell[c]{\#Avg.\\Documents\\per Query} \\

\midrule
SciFact (test)             &   300 &     5,183 & 100 & 4,131 &  1.13 \\
Natural Questions (test)   & 3,452 & 2,681,468 & 100 & 9,936 &  1.22 \\
MS MARCO (dev)             & 6,980 & 8,841,823 & 100 & 9,819 &  1.07 \\
NFCorpus (dev)             &   324 &     3,633 & 100 & 3,242 & 35.14 \\
HotpotQA (dev)             & 5,447 & 5,233,329 & 100 & 9,968 &  2.00 \\
FiQA (dev)                 &   500 &    57,638 & 100 & 8,349 &  2.48 \\
ArguAna (test)             & 1,406 &     8,674 & 100 & 5,110 &  1.00 \\
Quora Question Pairs (dev) & 5,000 &   522,931 & 100 & 9,891 &  1.53 \\
FEVER (dev)                & 6,666 & 5,416,568 & 100 & 9,731 &  1.21 \\
CLIMATE-FEVER (test)       & 1,535 & 5,416,593 & 100 & 7,696 &  3.05 \\
\bottomrule

\end{tabular}
}
\end{threeparttable}

\end{table}

%% file: tables/data_obscuration.tex
\begin{table*}[!htbp]

\centering

\caption{Similarity between shadow queries and corresponding documents.
}
\label{table:obscuration}

\renewcommand{\arraystretch}{.8}
\setlength{\tabcolsep}{6pt}

\begin{threeparttable}
\resizebox{0.8\linewidth}{!}{
\begin{tabular}{
@{\hspace{3pt}}l
ccc
c
c
c
p{0pt}
c
c
}
\toprule
\multicolumn{1}{c}{\multirow{3.5}{*}{Dataset}} &
\multicolumn{6}{c}{Embedding Decoupling} && 
\multicolumn{2}{c}{Semantic Decomposition} \\
\cmidrule(lr){2-7}
\cmidrule(lr){9-10}

&
\makecell[c]{BLEU ($\downarrow$)} &
\makecell[c]{R-1 ($\downarrow$)} & 
\makecell[c]{R-2 ($\downarrow$)} & 
\makecell[c]{R-L ($\downarrow$)} &
\makecell[c]{METEOR ($\downarrow$)} &
\makecell[c]{Emb. Sim. ($\uparrow$)} &&
\makecell[c]{Document\\Coverage ($\uparrow$)} &
\makecell[c]{Concat.\\Emb. Sim. ($\uparrow$)} \\
\midrule

SciFact       & 0.0096 & 0.2923 & 0.0983 & 0.2336 & 0.1616 & 0.7954 && 0.6812 & 0.8775 \\
NQ            & 0.0148 & 0.2983 & 0.1193 & 0.2380 & 0.1629 & 0.7960 && 0.6549 & 0.8852 \\
MS MARCO      & 0.0068 & 0.2606 & 0.0939 & 0.2171 & 0.1316 & 0.7755 && 0.6180 & 0.8647 \\
NFCorpus      & 0.0074 & 0.2679 & 0.0851 & 0.2145 & 0.1443 & 0.7958 && 0.6272 & 0.8744 \\
HotpotQA      & 0.0114 & 0.3070 & 0.1353 & 0.2549 & 0.1567 & 0.8046 && 0.6921 & 0.9162 \\
FiQA          & 0.0041 & 0.2045 & 0.0569 & 0.1647 & 0.1081 & 0.7496 && 0.4846 & 0.8293 \\
ArguAna       & 0.0023 & 0.1949 & 0.0489 & 0.1561 & 0.1015 & 0.7705 && 0.5381 & 0.8582 \\
Quora         & 0.0527 & 0.3341 & 0.1320 & 0.3039 & 0.2733 & 0.8310 && 0.7980 & 0.7841 \\
FEVER         & 0.0138 & 0.3145 & 0.1427 & 0.2629 & 0.1654 & 0.8053 && 0.7240 & 0.9146 \\
CLIMATE-FEVER & 0.0097 & 0.2815 & 0.1147 & 0.2340 & 0.1451 & 0.7938 && 0.7030 & 0.8944 \\
\bottomrule

\end{tabular}
}

\begin{tablenotes}
\footnotesize
\centering
\hspace*{1em}
\begin{minipage}{0.8\linewidth}
    \item[*] $\downarrow$ (or $\uparrow$) indicates that lower (greater) values are desired for these metrics.
\end{minipage}
\end{tablenotes}

\end{threeparttable}

\end{table*}

%% file: tables/utility.tex
\begin{table*}[!htbp]

\centering

\caption{
    Retrieval utility comparison between no defense, embedding noise, and \codename.
}
\label{table:utility}

\renewcommand{\arraystretch}{.8}
\setlength{\tabcolsep}{3pt}

\begin{threeparttable}
\resizebox{\linewidth}{!}{
\begin{tabular}{
l
rrrrr | p{0pt}
rrrrr p{1pt}
rrrrr
}
\toprule
\multicolumn{1}{c}{\multirow{2.5}{*}{Dataset}} &
\multicolumn{5}{c|}{No Defense (Baseline)} &&
\multicolumn{5}{c}{Embedding Noise} &&
\multicolumn{5}{c}{\codename} \\
\cmidrule(lr){2-6}
\cmidrule(lr){8-12}
\cmidrule(lr){14-18}

&
\makecell[c]{NDCG ($\uparrow$)} &
\makecell[c]{MAP ($\uparrow$)} & 
\makecell[c]{Recall ($\uparrow$)} & 
\makecell[c]{Prec.($\uparrow$)} &
\makecell[c]{Acc. ($\uparrow$)} && 

\makecell[c]{NDCG ($\uparrow$)} &
\makecell[c]{MAP ($\uparrow$)} & 
\makecell[c]{Recall ($\uparrow$)} & 
\makecell[c]{Prec. ($\uparrow$)} &
\makecell[c]{Acc. ($\uparrow$)} && 

\makecell[c]{NDCG ($\uparrow$)} &
\makecell[c]{MAP ($\uparrow$)} & 
\makecell[c]{Recall ($\uparrow$)} & 
\makecell[c]{Prec. ($\uparrow$)} &
\makecell[c]{Acc. ($\uparrow$)} \\
\midrule

SciFact & 
0.2428 & 0.2070 & 0.3467 & 0.0370 & 0.3500 &&
$-$0.0131 & $-$0.0163 & $+$0.0000 & $+$0.0000 & $+$0.0000 &&
\textbf{$+$0.0382} & \textbf{$+$0.0390} & \textbf{$+$0.0416} & \textbf{$+$0.0040} & \textbf{$+$0.0500} \\
NQ & 
0.4390 & 0.3859 & 0.5650 & 0.0660 & 0.6000 &&
$-$0.0018 & $-$0.0057 & $+$0.0100 & $+$0.0010 & $+$0.0100 &&
\textbf{$+$0.0091} & \textbf{$+$0.0043} & \textbf{$+$0.0317} & \textbf{$+$0.0030} & \textbf{$+$0.0400} \\
MS MARCO & 
0.4437 & 0.3838 & 0.6283 & 0.0670 & 0.6400 &&
\textbf{$-$0.0023} & \textbf{$-$0.0141} & \textbf{$+$0.0350} & \textbf{$+$0.0040} & \textbf{$+$0.0400} &&
$-$0.0302 & $-$0.0514 & \textbf{$+$0.0350} & $+$0.0030 & \textbf{$+$0.0400} \\
NFCorpus & 
0.2113 & 0.0705 & 0.1107 & 0.1550 & 0.5400 &&
$-$0.0057 & \textbf{$-$0.0023} & $-$0.0015 & $-$0.0040 & \textbf{$+$0.0100} &&
\textbf{$-$0.0024} & $-$0.0031 & \textbf{$+$0.0021} & \textbf{$+$0.0110} & $+$0.0000 \\
HotpotQA & 
0.5590 & 0.4568 & 0.6050 & 0.1210 & 0.8900 &&
\textbf{$-$0.0031} & \textbf{$-$0.0015} & $-$0.0100 & $-$0.0020 & $+$0.0000 &&
$-$0.0141 & $-$0.0149 & \textbf{$+$0.0050} & \textbf{$+$0.0010} & $-$0.0100 \\
FiQA & 
0.2610 & 0.1949 & 0.3120 & 0.0750 & 0.5000 &&
\textbf{$-$0.0024} & \textbf{$-$0.0062} & \textbf{$+$0.0158} & \textbf{$+$0.0030} & \textbf{$+$0.0200} &&
$-$0.0153 & $-$0.0141 & $-$0.0005 & $-$0.0060 & $+$0.0100 \\
ArguAna & 
0.4118 & 0.3337 & 0.6600 & 0.0660 & 0.6600 &&
\textbf{$-$0.0049} & \textbf{$-$0.0016} & \textbf{$-$0.0200} & \textbf{$-$0.0020} & \textbf{$-$0.0200} &&
$-$0.0747 & $-$0.0638 & $-$0.1100 & $-$0.0110 & $-$0.1100 \\
Quora & 
0.9146 & 0.8745 & 0.9822 & 0.1510 &	1.0000 &&
\textbf{$-$0.0015} & \textbf{$-$0.0020} & \textbf{$+$0.0033} & \textbf{$+$0.0010} & $+$0.0000 &&
$-$0.0622 & $-$0.0778 & $-$0.0100 & $-$0.0020 & $-$0.0100 \\
FEVER & 
0.6407 & 0.5881 & 0.7485 & 0.0850 & 0.8300 &&
\textbf{$+$0.0076} & \textbf{$+$0.0091} & $+$0.0017 & $+$0.0010 & $+$0.0000 &&
$-$0.0026 & $-$0.0258 & \textbf{$+$0.0640} & \textbf{$+$0.0080} & \textbf{$+$0.0600} \\
CLIMATE$-$FEVER & 
0.1876 & 0.1228 & 0.2468 & 0.0600 & 0.5300 &&
\textbf{$+$0.0015} & \textbf{$+$0.0016} & \textbf{$+$0.0054} & \textbf{$+$0.0020} & \textbf{$+$0.0100} &&
$-$0.0058 & $-$0.0009 & $-$0.0260 & $-$0.0040 & $-$0.0200 \\


\bottomrule
\end{tabular}
}
\begin{tablenotes}
    \footnotesize
    \item[*] $\uparrow$ indicates that greater values are desired for these metrics.
    \item[**] The ``$+$'' or ``$-$'' before the numbers indicates the increase or decrease in utility compared to no defense.
    \item[***] The bold numbers indicate the better retrieval utility between defense methods (embedding noise \textit{vs.} \codename).
\end{tablenotes}
\end{threeparttable}
\end{table*}

%% file: tables/privacy.tex
\begin{table*}[!htbp]

\centering

\caption{
    Defense efficacy comparison between no defense, embedding noise, and \codename.
}
\label{table:privacy}

\renewcommand{\arraystretch}{.8}
\setlength{\tabcolsep}{4pt}

\begin{threeparttable}
\resizebox{\linewidth}{!}{
\begin{tabular}{
l
rrrr@{\hspace{3pt}}r |
p{0pt}@{\hspace{0pt}}
rrrr@{\hspace{3pt}}r
p{0pt}@{\hspace{0pt}}
rrrr@{\hspace{3pt}}r
}
\toprule
\multicolumn{1}{c}{\multirow{2.5}{*}{Dataset}} &
\multicolumn{5}{c|}{No Defense (Baseline)} &&
\multicolumn{5}{c}{Embedding Noise} &&
\multicolumn{5}{c}{\codename}\\
\cmidrule(lr){2-6}
\cmidrule(lr){8-12}
\cmidrule(lr){13-18}

&
\makecell[c]{BLEU ($\downarrow$)} &
\makecell[c]{R-1 ($\downarrow$)} & 
\makecell[c]{R-2 ($\downarrow$)} & 
\makecell[c]{R-L ($\downarrow$)} &
\makecell[c]{METEOR ($\downarrow$)} && 

\makecell[c]{BLEU ($\downarrow$)} &
\makecell[c]{R-1 ($\downarrow$)} & 
\makecell[c]{R-2 ($\downarrow$)} & 
\makecell[c]{R-L ($\downarrow$)} &
\makecell[c]{METEOR ($\downarrow$)} && 

\makecell[c]{BLEU ($\downarrow$)} &
\makecell[c]{R-1 ($\downarrow$)} & 
\makecell[c]{R-2 ($\downarrow$)} & 
\makecell[c]{R-L ($\downarrow$)} &
\makecell[c]{METEOR ($\downarrow$)} \\
\midrule

SciFact & 
0.8015 & 0.9354 & 0.8265 & 0.8795 & 0.9157 &&
$-$0.7509 & $-$0.5723 & $-$0.7346 & $-$0.6317 & $-$0.6566 &&
\textbf{$-$0.7782} & \textbf{$-$0.6583} & \textbf{$-$0.7425} & \textbf{$-$0.6707} & \textbf{$-$0.7384} \\
NQ & 
0.8676 & 0.9664 & 0.8971 & 0.9331 & 0.9452 &&
$-$0.7813 & $-$0.5294 & $-$0.7465 & $-$0.6310 & $-$0.6226 &&
\textbf{$-$0.8324} & \textbf{$-$0.6514} & \textbf{$-$0.7814} & \textbf{$-$0.6906} & \textbf{$-$0.7662} \\
MS MARCO &
0.7263 & 0.9081 & 0.7758 & 0.8479 & 0.8743 &&
$-$0.6648 & $-$0.5103 & $-$0.6530 & $-$0.5710 & $-$0.5833 &&
\textbf{$-$0.7080} & \textbf{$-$0.6391} & \textbf{$-$0.6932} & \textbf{$-$0.6376} & \textbf{$-$0.7272} \\
NFCorpus & 
0.8279 & 0.9375 & 0.8494 & 0.8991 & 0.9288 &&
$-$0.7828 & $-$0.5722 & $-$0.7637 & $-$0.6503 & $-$0.6688 &&
\textbf{$-$0.8093} & \textbf{$-$0.6822} & \textbf{$-$0.7924} & \textbf{$-$0.7124} & \textbf{$-$0.7734} \\
HotpotQA &
0.9302 & 0.9794 & 0.9429 & 0.9646 & 0.9765 &&
$-$0.8581 & $-$0.5842 & $-$0.8182 & $-$0.6821 & $-$0.6828 &&
\textbf{$-$0.9059} & \textbf{$-$0.6767} & \textbf{$-$0.8207} & \textbf{$-$0.7235} & \textbf{$-$0.8168} \\
FiQA &
0.5966 & 0.8542 & 0.6649 & 0.7592 & 0.8059 &&
$-$0.5337 & $-$0.4619 & $-$0.5417 & $-$0.4986 & $-$0.5181 &&
\textbf{$-$0.5875} & \textbf{$-$0.6371} & \textbf{$-$0.6135} & \textbf{$-$0.5869} & \textbf{$-$0.6814} \\
ArguAna &
0.7708 & 0.9317 & 0.8106 & 0.8496 & 0.9131 &&
$-$0.7228 & $-$0.5264 & $-$0.7020 & $-$0.5933 & $-$0.6257 &&
\textbf{$-$0.7610} & \textbf{$-$0.7213} & \textbf{$-$0.7616} & \textbf{$-$0.6937} & \textbf{$-$0.7946} \\
Quora & 
0.3052 & 0.6854 & 0.5013 & 0.6282 & 0.6635 &&
$-$0.2653 & $-$0.3127 & $-$0.3728 & $-$0.3541 & $-$0.3036 &&
\textbf{$-$0.2686} & \textbf{$-$0.4176 }& \textbf{$-$0.4032} & \textbf{$-$0.3868} & \textbf{$-$0.4459} \\
FEVER & 
0.8672 & 0.9622 & 0.9123 & 0.9478 & 0.9468 &&
$-$0.7887 & $-$0.5536 & $-$0.7719 & $-$0.6634 & $-$0.6211 &&
\textbf{$-$0.8351} & \textbf{$-$0.6364} & \textbf{$-$0.7745} & \textbf{$-$0.6906} & \textbf{$-$0.7727} \\
CLIMATE$-$FEVER & 
0.8361 & 0.9582 & 0.8718 & 0.9215 & 0.9338 &&
$-$0.7601 & $-$0.5420 & $-$0.7444 & $-$0.6311 & $-$0.6327 &&
\textbf{$-$0.8149} & \textbf{$-$0.6881} & \textbf{$-$0.7861} & \textbf{$-$0.7097} & \textbf{$-$0.7787} \\


\bottomrule
\end{tabular}
}
\begin{tablenotes}
    \footnotesize
    \item[*] $\downarrow$ indicates that lower values are desired for these metrics.
    \item[**] The ``$+$'' or ``$-$'' before the numbers indicates the increase or decrease in privacy risk compared to no defense.
    \item[***] The bold numbers indicate the better defense efficacy between defense methods (embedding noise \textit{vs.} \codename).
\end{tablenotes}
\end{threeparttable}
\end{table*}

%% file: sections/5_related_work.tex
\section{Related Work}
This section delves into existing EIAs, commonly used defenses, and existing document/query generation techniques. 

\subsection{Existing EIAs}
Embeddings produced by pre-trained language models are increasingly recognized as vulnerable to privacy attacks. Recently, a technique called embedding inversion allows the adversary to reconstruct the original texts from observed embeddings, posing a significant data privacy threat.

\paragraph{Gradient-based method} Song et al.~\cite{song2020information} firstly investigate the information leakage in embedding models. They point out that brute-force methods, such as enumerating all possible sequences from the vocabulary, are computationally infeasible. To address this challenge, they propose an EIA that reconstructs text with gradient-based optimization. Starting from randomly initialized dummy token embeddings, this method iteratively updates the dummy token embeddings through gradient descent to minimize the distance between the resulting embedding and the target embedding, thereby recovering a large portion of the original text. However, this method is still computationally expensive as it requires backpropagation through the embedding model for each reconstruction.

\paragraph{Generation-based methods} Li et al.~\cite{li2023sentence} introduce Generative EIA (GEIA), which trains a GPT-2 model to predict original text given a target embedding. However, this attack focuses on the sentence level, posing less of a threat to a large-scale corpus. More recently, Morris et al.~\cite{morris2023text} propose a state-of-the-art EIA, vec2text, that trains a T5 model to decode text given a target embedding. This attack primarily focuses on vector databases and targets document-level embedding, posing a significant threat to modern dense retrievers. In this work, we primarily aim to address this threat in vector databases.

\subsection{Existing Defenses}
Existing countermeasures for EIAs typically focus on modifying the resulting embeddings to obscure the inversion process. Morris et al.~\cite{morris2023text} inject random noise into the embeddings, but only a minimal level of noise can be applied, struggling to achieve a utility-privacy balance. Zhuang et al.~\cite{zhuang2024understanding} propose a secret scaling method, where a scaling factor only known to the user is multiplied by the user query and document embeddings. However, this defense can be easily mitigated by a simple vector normalization, resulting in no effect in practice. Unlike these existing defenses, we propose a shadow query generation defense that shifts the defense paradigm from modifying document embeddings to replacing the document embeddings with shadow query embeddings, achieving semantic decoupling and semantic decomposition. Related research is also presented for vision models~\cite{wang2025catch, wang2026re, wang2025ai, wang2024corelocker} and for model verification~\cite{ma2025convex}.

\subsection{Document/Query Generation}
The generation of documents or queries is a technique primarily used to improve the retrieval performance or to augment the training datasets for dense retrievers. On the query side, Nogueira et al.~\cite{nogueira2019document} propose doc2query, which leverages a sequence-to-sequence model to generate a pseudo query given a document, and appends this pseudo query to the document to improve retrieval effectiveness. Later on, they propose docTTTTTquery~\cite{nogueira2019doc2query}, which uses a pre-trained T5 model to further boost the query generation quality. On the document side, Wang et al.~\cite{wang2023query2doc} introduce query2doc, which uses a pre-trained LLM to augment the document corpus given a query, to improve the training generalization. Unlike these methods that focus on the utility aspect of IR systems, \codename introduces a query generation pipeline that specifically focuses on enhancing the document data privacy of modern dense retrievers.

%% file: sections/6_conclusion.tex
\section{Conclusion}
In this work, we propose \codename, a novel defense mechanism against EIAs in vector databases by shadow query generation. \codename is built upon an important insight: EIAs fundamentally rely on the strong coupling between embeddings and their underlying textual content. By replacing the document embeddings with the shadow query embeddings in the vector database, \codename achieves strong privacy protection by breaking the inherent connection between the stored embeddings and the underlying texts (embedding decoupling), while facilitating an effective document retrieval by preserving important semantic facets of the original documents in the generated shadow queries (semantic decomposition). \codename demonstrates a paradigm shift from directly modifying the document embeddings to enabling semantic decomposition and embedding decoupling for effective mitigation against EIAs in vector databases.

%% file: sections/Appendix.tex
\appendix

\subsection{Shadow Query Generation Prompt Template}
\label{prompt_template}

\definecolor{skyblue}{rgb}{0.529, 0.808, 0.922}
\begin{tcolorbox}[
    fontupper=\small,
    colback=white,    
    colframe=skyblue!40!white, 
    width=\columnwidth,       
    sharp corners,           
    title=Shadow query generation prompt template, 
    fonttitle=\bfseries,     
    coltitle=black,           
    left=0mm, right=0mm, top=1mm, bottom=1mm,
]

You are an expert in information retrieval. Below is a document passage from a database. Your task is to generate \{$n_g$\} diverse queries that would retrieve this document. Each query should be concise, natural, and reflect a different perspective or way a user might search for this content. Queries may be questions, statements, or keyword-style phrases. Avoid semantic repetition. Do not repeat the same query, or generate queries with the same semantics.
\\[1ex]
Document: \{$d$\}
\\[1ex]
Keep your thinking process short. After your thinking process, start a new line and output a list of queries as a valid Python list of strings, clearly after the 'Queries:' prefix below. Do not include explanations or any other text after your final output. Now, let's try this step by step!
\end{tcolorbox}

\subsection{Generated Shadow Queries}
\label{generated_shadow_queries}

An example of the generated shadow queries is shown below. A concise set of shadow queries is generated, given the target document. Notably, most of the content in the target document is covered by the shadow queries, demonstrating semantic decomposition. These shadow queries are either in a compact fact format or in a question format, resembling plausible user queries at retrieval time. 

\definecolor{skyblue}{rgb}{0.529, 0.808, 0.922}
\begin{tcolorbox}[
    fontupper=\small,
    colback=white,    
    colframe=skyblue!40!white, 
    width=\columnwidth,       
    sharp corners,           
    title=Example of generated shadow queries, 
    fonttitle=\bfseries,     
    coltitle=black,           
    left=0mm, right=0mm, top=1mm, bottom=1mm,
    breakable
]

\textbf{Document}: \\[1ex]
BACKGROUND Increased expression of the tetraspanin TSPAN7 has been observed in a number of cancers; however, it is unclear how TSPAN7 plays a role in cancer progression. METHODS \uline{We investigated the expression of TSPAN7 in the hematological malignancy multiple myeloma (MM)}$^\text{\ding{179}}$ and \uline{assessed the consequences of TSPAN7 expression in the adhesion, migration and growth of MM plasma cells (PC) in vitro and in bone marrow (BM) homing and tumor growth in vivo.}$^\text{\ding{174}}$ Finally, we characterized the association of TSPAN7 with cell surface partner molecules in vitro. RESULTS \uline{TSPAN7 was found to be highly expressed at the RNA and protein level in CD138(+) MM PC from approximately 50\% of MM patients.}$^\text{\ding{172}}$ \uline{TSPAN7 overexpression in the murine myeloma cell line 5TGM1 significantly reduced tumor burden in 5TGM1/KaLwRij mice 4 weeks after intravenous administration of 5TGM1 cells.}$^\text{\ding{173}, \ding{180}}$ \uline{While TSPAN7 overexpression did not affect cell proliferation in vitro,}$^\text{\ding{178}}$ \uline{TSPAN7 increased 5TGM1 cell adhesion to BM stromal cells and transendothelial migration.}$^\text{\ding{175}}$ \uline{In addition, TSPAN7 was found to associate with the molecular chaperone calnexin on the cell surface.}$^\text{\ding{176}, \ding{181}}$ CONCLUSION \uline{These results suggest that elevated TSPAN7 may be associated with better outcomes for up to 50\% of MM patients.}$^\text{\ding{177}}$
\\[1ex]
\textbf{Shadow queries}:\\[1ex]
\ding{172} TSPAN7 expression in multiple myeloma plasma cells.\\[0ex]
\ding{173} Does TSPAN7 overexpression reduce tumor burden in multiple myeloma?\\[0ex]
\ding{174} Role of TSPAN7 in bone marrow homing of myeloma cells.\\[0ex]
\ding{175} TSPAN7 effects on cell adhesion and migration in MM.\\[0ex]
\ding{176} Association between TSPAN7 and calnexin in cancer cells.\\[0ex]
\ding{177} Impact of TSPAN7 on multiple myeloma patient prognosis.\\[0ex]
\ding{178} In vitro studies of TSPAN7 and myeloma cell proliferation.\\[0ex]
\ding{179} TSPAN7 expression levels in hematological malignancies.\\[0ex]
\ding{180} 5TGM1 cell line and TSPAN7 overexpression outcomes.\\[0ex]
\ding{181} Molecular chaperone calnexin interaction with TSPAN7.

\end{tcolorbox}

\subsection{Empirical Implementation}
\label{appendix: implementation}

\paragraph{Experiment enviornment} We implement \codename in Python 3.12.3 with PyTorch 2.3.0 and transformers 4.44.2. All experiments are conducted on a high-performance computing cluster running Rocky Linux equipped with NVIDIA H100 GPUs.

\paragraph{Random factors} All experiments are run with the same random seed to minimize the effect of random factors, such as the K-Means initialization and scattering indexing, on the experiment results.

\subsection{Comprehensive Details of Evaluated Datasets}
\label{appendix: dataset details}

The comprehensive details of the evaluated datasets are listed as follows:

\begin{itemize}[left=0pt, nosep]
    \item \textbf{SciFact}~\cite{wadden2020fact}: A dataset for scientific fact verification. 
    Each query is a scientific claim, and the documents are abstracts from biomedical research papers. 
    \item \textbf{Natural Questions}~\cite{kwiatkowski2019natural}: A dataset for open-domain question answering. 
    Each query is a real user question issued to the Google search engine, and the documents are Wikipedia articles.
    \item \textbf{MS MARCO}~\cite{nguyen2016ms}: A dataset for information retrieval and machine reading comprehension.
    Each query is a real user query issued to the Bing search engine, and the documents are passages extracted from web pages.
    \item \textbf{NFCorpus}~\cite{boteva2016full}: A dataset for biomedical information retrieval. 
    Each query is a non-technical query gathered from the NutritionFacts.org~\cite{nutritionfacts2025} site, and the documents are relevance judgments extracted from medical documents.
    \item \textbf{HotpotQA}~\cite{yang2018hotpotqa}: A dataset for question answering. 
    Each query is a question, and the documents are paragraphs from Wikipedia articles.
    \item \textbf{FiQA}~\cite{maia201818}: A dataset for financial-domain information retrieval. 
    Each query is a real-world, finance-related question, and the documents are passages from varied sources such as microblogs, news, and reports.
    \item \textbf{ArguAna}~\cite{wachsmuth2018retrieval}: A dataset for counter-argument retrieval. 
    Each query is an argumentative claim, e.g., an argument, and the documents are candidate counterarguments sourced from the internet. 
    \item \textbf{Quora Question Pairs}~\cite{iyer2017quoraquestionpairs}: A dataset for paraphrase identification and retrieval. 
    Each query is a question, and the document is another question. The task is to determine whether the two questions are semantically equivalent.
    \item \textbf{FEVER}~\cite{thorne2018fever}: A dataset for fact extraction and verification. 
    Each query is a claim generated by mutating sentences from Wikipedia, and the documents are Wikipedia pages. 
    \item \textbf{CLIMATE-FEVER}~\cite{diggelmann2020climate}: A dataset for fact verification tailored to climate change claims, 
    following the style of the FEVER dataset.
\end{itemize}

\subsection{Excluded Datasets from Evaluation}
\label{appendix: removed_datasets}
The statistical details and the reasons for exclusion of the datasets from the BEIR benchmark are summarized in Table~\ref{table:removed_datasets}. 

\input{tables/removed_datasets}

\subsection{Hyperparameter Search}
\label{hyper}
Considering the limited computational resources available, we begin the hyperparameter search for \codename by generating $n_g = 10$ shadow queries per sampled document. K-Means clustering is then applied to group them, and then one query from each cluster is randomly selected to form the final set. We experiment with $n_k \in \{1, 5, 10\}$ and find that $n_k=10$ yields the best trade-off between utility and privacy. Further manual inspection confirms that the generated shadow queries are already diverse and effectively capture different semantic aspects of the document, attributed to both the prompt design and the capacity of the employed generative language model. Therefore, without loss of generality of applying the clustering algorithm, we adopt $n_k=10$ throughout the experiments. Given more computational resources, increasing the number of initially generated shadow queries is anticipated to improve the retrieval performance.

To ensure a fair comparison with baseline defenses, we also run a hyperparameter search for $\sigma$, the noise level in the embedding noise defense. We experiment with $\sigma \in \{0.001, 0.01, 0.2\}$, and observe that $\sigma = 0.01$ yields comparable utility with $n_k=10$ in \codename.
Larger values of $\sigma$ further degrade the retrieval utility, consistent with prior findings by Morris et al.~\cite{morris2023text} and Zhuang et al.~\cite{zhuang2024understanding}, where $\sigma = 0.01$ is also suggested as a reasonable noise level for embedding noise.

The details of the hyperparameter search are presented in Table~\ref{table:hyper}.

\input{tables/hyper}

\subsection{Ablation Studies}
\label{appendix:ablation}
We present the ablation studies.
In Section~\ref{ablation: truncation}, we lift the document truncation constraint, showcasing \codename's generality on variable-length documents. In Section~\ref{ablation: embedding}, we investigate the retrieval utility with a different embedding model, demonstrating \codename's model-agnostic ability. 
As the attack model we adopt only supports truncated documents and a fixed embedding model, we do not investigate the attack efficacy in these ablation studies.

\subsection{Retrieval Utility on Untruncated Documents}
\label{ablation: truncation}
In the main experiments, all documents are truncated to 32 tokens to align with the constraints of the vec2text attack model, which is trained to reconstruct texts with a fixed length of 32 tokens. We also adopt this setting for \codename, where a truncated document is used for the shadow query generation, to enable a fair evaluation. However, this truncation of the documents limits the semantic richness available for encoding the documents or the shadow query generation, potentially affecting the retrieval performance. To isolate the effect of this constraint, we conduct an ablation study using full-length documents to evaluate the resulting impact on retrieval utility. 
This setup closely aligns with a practical production scenario, where the documents are in full length. 
We perform this analysis on the SciFact dataset and only evaluate the retrieval utility, as the attack model cannot handle variable-length inputs, and thus the defense efficacy cannot be assessed in this setting.

\paragraph{Results} The results are presented in Table~\ref{table:ablation_no_truncate}. Across all defense scenarios, we observe substantial improvements in retrieval utility when full-length documents are used. Notably, \codename exhibits the largest performance gains: $+0.3684$ in NDCG, $+0.3556$ in MAP, $+0.3934$ in Recall, $+0.0430$ in Precision, and $+0.4000$ in Accuracy, compared to its truncated counterpart. It outperforms both the undefended setting and the embedding noise baseline. These findings demonstrate that more comprehensive document content enables the generation of higher-quality shadow queries, thereby improving retrieval effectiveness. 

\input{tables/ablation_no_truncate}
\input{tables/ablation_embed}

\subsection{Retrieval Utility on Different Embedding Model}
\label{ablation: embedding}
In the main experiments, the GTR-T5 model~\cite{ni2021large} is employed as the embedding model to align with the attack model provided in vec2text, which is trained on the embeddings generated by it. To assess the effectiveness of \codename on a different embedding model, we conduct an ablation study that employs the DPR~\cite{karpukhin-etal-2020-dense} as the embedding model. Notably, the GTR-T5 model is trained with the cosine similarity objective, and the DPR model is trained with the dot product similarity objective. This further assesses \codename's model-agnostic ability.

\paragraph{Results} The results are presented in Table~\ref{table:ablation_embed}. We observe that for all three scenarios, the retrieval performance experiences a downfall compared to the GTR-T5 model, mainly due to the poorer semantic embedding ability of the DPR model. Despite the performance downfall, all three scenarios remain retrieval performance on par with each other, confirming that \codename is a model-agnostic defense that can directly be integrated into general dense retrievers.





\subsection{Adaptive attack}
\label{appendix: adaptive}
\paragraph{Adversary capabilities} In this scenario, the attacker has the same objective, which is to reconstruct the original documents by observing the embeddings in the vector database. Besides the capabilities outlined in the threat model in Section~\ref{sec:threat_model}, the attacker has complete knowledge of the defense mechanism, including details such as $n_k$ shadow queries for each document, and has access to the LLM employed for the shadow query generation. More importantly, the attacker knows that the embeddings observed in the vector database are of the shadow queries for the corresponding target documents.

\paragraph{Attack strategy} To exploit this extra knowledge, the attacker first aims to identify the $n_k$ embeddings of the shadow queries for the same document and reconstruct these shadow queries one by one. From these reconstructed shadow queries, the attacker aims to piece together the original documents. 
Recall that in the indexing step, \codename scatters the embeddings of the shadow queries for each document in the vector database. If they are inserted in a sequential order, the attacker can easily identify all of them by locating the first embedding. Knowing the scattering approach, for an observed embedding, the attacker attempts to locate the other shadow query embeddings by computing the cosine similarities between the observed embedding and all other embeddings, and locates the $n_k - 1$ embeddings that exhibit the highest similarities. 
After these shadow queries are reconstructed, the attacker aims to piece together the target document from the shadow queries by using the same LLM for the shadow query generation. Specifically, a prompt template is used to guide the LLM to attempt to reconstruct the original document from the identified shadow queries. The complete prompt template is shown below.

\definecolor{skyblue}{rgb}{0.529, 0.808, 0.922}
\begin{tcolorbox}[
    fontupper=\small,
    colback=white,    
    colframe=skyblue!40!white, 
    width=\columnwidth,       
    sharp corners,           
    title=Original document reconstruction prompt template, 
    fonttitle=\bfseries,     
    coltitle=black,           
    left=0mm, right=0mm, top=1mm, bottom=1mm,
]

You are an expert in information retrieval. Below is a list of possible shadow queries generated from a document passage. Your task is to piece together the original document passage from these queries. Generate a coherent and comprehensive document passage that would be relevant to these queries.
\\[1ex]
Shadow Queries: \{$S$\}
\\[1ex]
Keep your thinking process short. After your thinking process, start a new line and output the document as a string, clearly after the 'Document:' prefix below. Do not include explanations or any other text after your final output. Now, let's try this step by step!

\end{tcolorbox}

\input{tables/adaptive}

\paragraph{Results} The results on the adaptive attack are presented in Table~\ref{table:adaptive}. Although the adaptive attack demonstrates increased attack efficacy, the overall data recovery remains relatively low, with BLEU score ranging from $0.0151$ to $0.1446$, R-1 from $0.2083$ to $0.4517$, R-2 from $0.0511$ to $0.2192$, R-L from $0.1665$ to $0.3530$, and METEOR from $0.1822$ to $0.3807$. This is due to the set of shadow queries generated from the same target document does not always exhibit similar semantics, as they are designed to cover distinct semantic facets of the documents in different query perspectives. Compared to the attack success rate on an undefended system or one with embedding noise applied, \codename effectively withstands even challenging adaptive attack and exhibits exceptional robustness, where a powerful adversary possesses complete knowledge of the applied defense.

%% file: tables/removed_datasets.tex
\begin{table}[!htp]

\centering

\caption{Data split and statistical details of the excluded datasets.}
\label{table:removed_datasets}

\renewcommand{\arraystretch}{.9}
\setlength{\tabcolsep}{1pt}

\resizebox{\linewidth}{!}{
\begin{threeparttable}
\begin{tabular}{l rrrrr}

\toprule
\makecell[c]{Dataset} &
\makecell[c]{\#Original\\Queries} &
\makecell[c]{\#Original\\Documents} &
\makecell[c]{\#Sampled\\Queries} &
\makecell[c]{\#Sampled\\Documents} &
\makecell[c]{\#Avg.\\Documents\\per Query} \\

\midrule
TREC-NEWS\tnote{1,2}   &    50 &   171,332 &  50 & 36,434 & 493.50 \\
Touche-2020\tnote{1,2} &    49 &   382,545 &  49 &  6,310 &  19.00 \\
CQADupstack\tnote{3}                  & -     & -         & -   & -      & -      \\
DBPedia\tnote{1,2}     &   400 & 4,635,922 & 100 & 18,035 &  38.20 \\
SCIDOCS\tnote{2}       & 1,000 &    25,657 & 100 & 10,050 &   4.90 \\
\bottomrule

\end{tabular}
\begin{tablenotes}
    \footnotesize
    \item[1] Large number of relevant documents per query.
    \item[2] Non-binary scale of relevant scores. 
    \item[3] Not a standard IR dataset.
\end{tablenotes}
\end{threeparttable}
}
\end{table}

%% file: tables/hyper.tex
\begin{table*}[!htbp]

\centering

\caption{Hyperparameter search for \#shadow queries ($n_k$) and noise standard deviation ($\sigma$) on the subsampled SciFact dataset.}
\label{table:hyper}

\renewcommand{\arraystretch}{.8}
\setlength{\tabcolsep}{6pt}

\begin{threeparttable}
\resizebox{0.66\linewidth}{!}{
\begin{tabular}{
l
ccccc
p{1pt}
ccccc
}

\toprule
\makecell[c]{\multirow{2}{*}{Parameter}} &
\multicolumn{5}{c}{Utility} &&
\multicolumn{5}{c}{Privacy} \\

\cmidrule(lr){2-6}
\cmidrule(lr){8-12}

&
\makecell[c]{NDCG ($\uparrow$)} &
\makecell[c]{MAP ($\uparrow$)} & 
\makecell[c]{Recall ($\uparrow$)} & 
\makecell[c]{Prec. ($\uparrow$)} &
\makecell[c]{Acc. ($\uparrow$)} && 
\makecell[c]{BLEU ($\downarrow$)} & 
\makecell[c]{R-1 ($\downarrow$)} & 
\makecell[c]{R-2 ($\downarrow$)} & 
\makecell[c]{R-L ($\downarrow$)} & 
\makecell[c]{METEOR ($\downarrow$)} \\
\midrule

&
\multicolumn{11}{c}{\codename with different \#shadow queries ($n_k$)} \\
\midrule

$n_k=0$ & 
0.2428 & 0.2070 & 0.3467 & 0.0370 & 0.3500 &&
0.8015 & 0.9354 & 0.8265 & 0.8795 & 0.9157 \\

$n_k=1$ &
0.1740 & 0.1439 & 0.2650 & 0.0280 & 0.2700 &&
0.0312 & 0.2744 & 0.0739 & 0.2054 & 0.1724 \\

$n_k=5$ & 
0.2338 & 0.1966 & 0.3483 & 0.0370 & 0.3600 &&
0.0263 & 0.2865 & 0.0761 & 0.2160 & 0.1802 \\

\rowcolor{gray!20}
$n_k=\textbf{10}$ &
0.2810 & 0.2460 & 0.3883 & 0.0410 & 0.4000 &&
0.0233 & 0.2771 & 0.0840 & 0.2088 & 0.1773 \\

\midrule

&
\multicolumn{11}{c}{Embedding noise with different noise standard deviation ($\sigma$)} \\
\midrule

$\sigma=0$ & 
0.2428 & 0.2070 & 0.3467 & 0.0370 & 0.3500 &&
0.8015 & 0.9354 & 0.8265 & 0.8795 & 0.9157 \\

$\sigma=0.001$ &
0.2418 & 0.2058 & 0.3467 & 0.0370 & 0.3500 &&
0.5502 & 0.8467 & 0.6124 & 0.7443 & 0.8063 \\

\rowcolor{gray!20}
$\sigma=\textbf{0.01}$ &
0.2297 & 0.1907 & 0.3467 & 0.0370 & 0.3500 &&
0.0506 & 0.3631 & 0.0919 & 0.2478 & 0.2591 \\

$\sigma=0.02$ & 
0.2062 & 0.1712 & 0.3117 & 0.0330 & 0.3200 &&
0.0145 & 0.2198 & 0.0311 & 0.1556 & 0.1492 \\
\bottomrule

\end{tabular}
}
\begin{tablenotes}
\footnotesize
\centering
\hspace*{1em}
\begin{minipage}{0.66\linewidth}
    \item[*] $\downarrow$ (or $\uparrow$) indicates that lower (greater) values are desired for these metrics.
    \item[**] The chosen hyperparameter setting is highlighted in bold with a gray background.
\end{minipage}
\end{tablenotes}

\end{threeparttable}

\end{table*}

%% file: tables/ablation_no_truncate.tex
\begin{table*}[!htbp]

\centering

\caption{Retrieval utility on the untruncated SciFact dataset.}
\label{table:ablation_no_truncate}

\renewcommand{\arraystretch}{.8}
\setlength{\tabcolsep}{3pt}

\begin{threeparttable}
\resizebox{\linewidth}{!}{
\begin{tabular}{
l
rrrrr | p{1pt}
rrrrr p{1pt}
rrrrr p{1pt}
}
\toprule
\multicolumn{1}{c}{\multirow{2.5}{*}{Dataset}} &
\multicolumn{5}{c|}{No Defense (Baseline)} &&
\multicolumn{5}{c}{Embedding Noise} &&
\multicolumn{5}{c}{\codename}\\
\cmidrule(lr){2-6}
\cmidrule(lr){8-12}
\cmidrule(lr){14-18}

&
\makecell[c]{NDCG ($\uparrow$)} &
\makecell[c]{MAP ($\uparrow$)} & 
\makecell[c]{Recall ($\uparrow$)} & 
\makecell[c]{Prec. ($\uparrow$)} &
\makecell[c]{Acc. ($\uparrow$)} && 

\makecell[c]{NDCG ($\uparrow$)} &
\makecell[c]{MAP ($\uparrow$)} & 
\makecell[c]{Recall ($\uparrow$)} & 
\makecell[c]{Prec. ($\uparrow$)} &
\makecell[c]{Acc. ($\uparrow$)} && 

\makecell[c]{NDCG ($\uparrow$)} &
\makecell[c]{MAP ($\uparrow$)} & 
\makecell[c]{Recall ($\uparrow$)} & 
\makecell[c]{Prec. ($\uparrow$)} &
\makecell[c]{Acc. ($\uparrow$)} \\
\midrule

SciFact (truncated) & 
0.2428 & 0.2070 & 0.3467 & 0.0370 & 0.3500 &&
$-$0.0131 & $-$0.0163 & $+$0.0000 & $+$0.0000 & $+$0.0000 &&
\textbf{$+$0.0382} & \textbf{$+$0.0390} & \textbf{$+$0.0416} & \textbf{$+$0.0040} & \textbf{$+$0.0500} \\

SciFact (non-truncated) & 
0.5916 & 0.5492 & 0.7117 & 0.0740 & 0.7300 &&
$-$0.0104 & $-$0.0119 & $-$0.0050 & $-$0.0010 & $-$0.0100 &&
\textbf{$+$0.0578} & \textbf{$+$0.0524} & \textbf{$+$0.0700} & \textbf{$+$0.0100} & \textbf{$+$0.0700} \\



\bottomrule
\end{tabular}
}
\begin{tablenotes}
    \footnotesize
    \item[*] $\uparrow$ indicates that greater values are desired for these metrics.
    \item[**] The ``$+$'' or ``$-$'' before the numbers indicates the increase or decrease in utility compared to no defense.
    \item[***] The bold numbers indicate the better retrieval utility between defense methods (embedding noise \textit{vs.} \codename).
\end{tablenotes}
\end{threeparttable}
\end{table*}

%% file: tables/ablation_embed.tex
\begin{table*}[!htbp]

\centering

\caption{Retrieval utility on SciFact dataset with DPR model.}
\label{table:ablation_embed}

\renewcommand{\arraystretch}{.8}
\setlength{\tabcolsep}{3pt}

\begin{threeparttable}
\resizebox{\linewidth}{!}{
\begin{tabular}{
l
rrrrr | p{1pt}
rrrrr p{1pt}
rrrrr p{1pt}
}
\toprule
\multicolumn{1}{c}{\multirow{2.5}{*}{Dataset}} &
\multicolumn{5}{c|}{No Defense (Baseline)} &&
\multicolumn{5}{c}{Embedding Noise} &&
\multicolumn{5}{c}{\codename}\\
\cmidrule(lr){2-6}
\cmidrule(lr){8-12}
\cmidrule(lr){14-18}

&
\makecell[c]{NDCG ($\uparrow$)} &
\makecell[c]{MAP ($\uparrow$)} & 
\makecell[c]{Recall ($\uparrow$)} & 
\makecell[c]{Prec. ($\uparrow$)} &
\makecell[c]{Acc. ($\uparrow$)} && 

\makecell[c]{NDCG ($\uparrow$)} &
\makecell[c]{MAP ($\uparrow$)} & 
\makecell[c]{Recall ($\uparrow$)} & 
\makecell[c]{Prec. ($\uparrow$)} &
\makecell[c]{Acc. ($\uparrow$)} && 

\makecell[c]{NDCG ($\uparrow$)} &
\makecell[c]{MAP ($\uparrow$)} & 
\makecell[c]{Recall ($\uparrow$)} & 
\makecell[c]{Prec. ($\uparrow$)} &
\makecell[c]{Acc. ($\uparrow$)} \\
\midrule

SciFact (GTR-T5) & 
0.2428 & 0.2070 & 0.3467 & 0.0370 & 0.3500 &&
$-$0.0131 & $-$0.0163 & $+$0.0000 & $+$0.0000 & $+$0.0000 &&
\textbf{$+$0.0382} & \textbf{$+$0.0390} & \textbf{$+$0.0416} & \textbf{$+$0.0040} & \textbf{$+$0.0500} \\

SciFact (DPR) & 
0.0897 & 0.0702 & 0.1483 & 0.0170 & 0.1600 &&
$+$0.0229 & $+$0.0269 & $+$0.0100 & $+$0.0010 & $+$0.0100 &&
\textbf{$+$0.0441} & \textbf{$+$0.0478} & \textbf{$+$0.0367} & \textbf{$+$0.0030} & \textbf{$+$0.0300} \\

\bottomrule
\end{tabular}
}
\begin{tablenotes}
    \footnotesize
    \item[*] $\downarrow$ indicates that lower values are desired for these metrics.
    \item[**] The ``$+$'' or ``$-$'' before the numbers indicates the increase or decrease in utility compared to no defense.
    \item[***] The bold numbers indicate the better retrieval utility between defense methods (embedding noise \textit{vs.} \codename).
\end{tablenotes}
\end{threeparttable}
\end{table*}

%% file: tables/adaptive.tex
\begin{table}[!htbp]

\centering

\caption{\codename defense efficacy on adaptive attack.}
\label{table:adaptive}

\renewcommand{\arraystretch}{.9}
\setlength{\tabcolsep}{6pt}

\begin{threeparttable}
\resizebox{\linewidth}{!}{
\begin{tabular}{
l
ccccc
}
\toprule
\multicolumn{1}{c}{\multirow{2.5}{*}{Dataset}} &
\multicolumn{5}{c}{\codename}\\
\cmidrule(lr){2-6}

&
\makecell[c]{BLEU ($\downarrow$)} &
\makecell[c]{R-1 ($\downarrow$)} & 
\makecell[c]{R-2 ($\downarrow$)} & 
\makecell[c]{R-L ($\downarrow$)} &
\makecell[c]{METEOR ($\downarrow$)} \\
\midrule

SciFact       & 0.0985 & 0.3514 & 0.1479 & 0.2715 & 0.3126 \\
NQ            & 0.1089 & 0.4004 & 0.1695 & 0.2988 & 0.3163 \\
MS MARCO      & 0.0636 & 0.3143 & 0.1153 & 0.2387 & 0.2455 \\
NFCorpus      & 0.0659 & 0.3137 & 0.1013 & 0.2290 & 0.2604 \\
HotpotQA      & 0.1446 & 0.4517 & 0.2192 & 0.3530 & 0.3807 \\
FiQA          & 0.0364 & 0.2409 & 0.0658 & 0.1804 & 0.1822 \\
ArguAna       & 0.0151 & 0.2563 & 0.0511 & 0.1861 & 0.1845 \\
Quora         & 0.0211 & 0.2083 & 0.0657 & 0.1665 & 0.2097 \\
FEVER         & 0.1286 & 0.4165 & 0.2046 & 0.3289 & 0.3566 \\
CLIMATE-FEVER & 0.1324 & 0.4224 & 0.1988 & 0.3258 & 0.3490 \\

\bottomrule
\end{tabular}
}
\begin{tablenotes}
    \footnotesize
    \item[*] $\downarrow$ indicates that lower values are desired for these metrics.
\end{tablenotes}
\end{threeparttable}
\end{table}

%% file: references.bib
@inproceedings{karpukhin-etal-2020-dense,
    title = "Dense Passage Retrieval for Open-Domain Question Answering",
    author = "Karpukhin, Vladimir  and
      Oguz, Barlas  and
      Min, Sewon  and
      Lewis, Patrick  and
      Wu, Ledell  and
      Edunov, Sergey  and
      Chen, Danqi  and
      Yih, Wen-tau",
    editor = "Webber, Bonnie  and
      Cohn, Trevor  and
      He, Yulan  and
      Liu, Yang",
    booktitle = "Proceedings of the 2020 Conference on Empirical Methods in Natural Language Processing (EMNLP)",
    month = nov,
    year = "2020",
    address = "Online",
    publisher = "Association for Computational Linguistics",
    url = "https://aclanthology.org/2020.emnlp-main.550/",
    doi = "10.18653/v1/2020.emnlp-main.550",
    pages = "6769--6781"
}

@article{lewis2020retrieval,
  title={Retrieval-augmented generation for knowledge-intensive nlp tasks},
  author={Lewis, Patrick and Perez, Ethan and Piktus, Aleksandra and Petroni, Fabio and Karpukhin, Vladimir and Goyal, Naman and K{\"u}ttler, Heinrich and Lewis, Mike and Yih, Wen-tau and Rockt{\"a}schel, Tim and others},
  journal={Advances in neural information processing systems},
  volume={33},
  pages={9459--9474},
  year={2020}
}

@inproceedings{song2020information,
  title={Information leakage in embedding models},
  author={Song, Congzheng and Raghunathan, Ananth},
  booktitle={Proceedings of the 2020 ACM SIGSAC conference on computer and communications security},
  pages={377--390},
  year={2020}
}

@article{morris2023text,
  title={Text embeddings reveal (almost) as much as text},
  author={Morris, John X and Kuleshov, Volodymyr and Shmatikov, Vitaly and Rush, Alexander M},
  journal={arXiv preprint arXiv:2310.06816},
  year={2023}
}

@inproceedings{zhuang2024understanding,
  title={Understanding and mitigating the threat of vec2text to dense retrieval systems},
  author={Zhuang, Shengyao and Koopman, Bevan and Chu, Xiaoran and Zuccon, Guido},
  booktitle={Proceedings of the 2024 Annual International ACM SIGIR Conference on Research and Development in Information Retrieval in the Asia Pacific Region},
  pages={259--268},
  year={2024}
}

@article{singhal2001modern,
  title={Modern information retrieval: A brief overview},
  author={Singhal, Amit and others},
  journal={IEEE Data Eng. Bull.},
  volume={24},
  number={4},
  pages={35--43},
  year={2001}
}

@inproceedings{ramos2003using,
  title={Using tf-idf to determine word relevance in document queries},
  author={Ramos, Juan and others},
  booktitle={Proceedings of the first instructional conference on machine learning},
  volume={242},
  number={1},
  pages={29--48},
  year={2003},
  organization={Citeseer}
}

@article{robertson2009probabilistic,
  title={The probabilistic relevance framework: BM25 and beyond},
  author={Robertson, Stephen and Zaragoza, Hugo and others},
  journal={Foundations and Trends{\textregistered} in Information Retrieval},
  volume={3},
  number={4},
  pages={333--389},
  year={2009},
  publisher={Now Publishers, Inc.}
}

@article{ni2021large,
  title={Large dual encoders are generalizable retrievers},
  author={Ni, Jianmo and Qu, Chen and Lu, Jing and Dai, Zhuyun and {\'A}brego, Gustavo Hern{\'a}ndez and Ma, Ji and Zhao, Vincent Y and Luan, Yi and Hall, Keith B and Chang, Ming-Wei and others},
  journal={arXiv preprint arXiv:2112.07899},
  year={2021}
}

@article{bert,
  title={Bert: Pre-training of deep bidirectional transformers for language understanding},
  author={Devlin, Jacob and Chang, Ming-Wei and Lee, Kenton and Toutanova, Kristina},
  journal={arXiv preprint arXiv:1810.04805},
  year={2018}
}

@article{roberta,
  title={Roberta: A robustly optimized bert pretraining approach},
  author={Liu, Yinhan and Ott, Myle and Goyal, Naman and Du, Jingfei and Joshi, Mandar and Chen, Danqi and Levy, Omer and Lewis, Mike and Zettlemoyer, Luke and Stoyanov, Veselin},
  journal={arXiv preprint arXiv:1907.11692},
  year={2019}
}

@article{touvron2023llama,
  title={Llama: Open and efficient foundation language models},
  author={Touvron, Hugo and Lavril, Thibaut and Izacard, Gautier and Martinet, Xavier and Lachaux, Marie-Anne and Lacroix, Timoth{\'e}e and Rozi{\`e}re, Baptiste and Goyal, Naman and Hambro, Eric and Azhar, Faisal and others},
  journal={arXiv preprint arXiv:2302.13971},
  year={2023}
}

@inproceedings{transformer,
  title={Attention is all you need},
  author={Vaswani, Ashish and Shazeer, Noam and Parmar, Niki and Uszkoreit, Jakob and Jones, Llion and Gomez, Aidan N and Kaiser, {\L}ukasz and Polosukhin, Illia},
  booktitle={the 2017 Advances in Neural Information Processing Systems (NeurIPS)},
  year={2017}
}

@article{zhang2010understanding,
  title={Understanding bag-of-words model: a statistical framework},
  author={Zhang, Yin and Jin, Rong and Zhou, Zhi-Hua},
  journal={International journal of machine learning and cybernetics},
  volume={1},
  pages={43--52},
  year={2010},
  publisher={Springer}
}

@article{shibata1999byte,
  title={Byte pair encoding: A text compression scheme that accelerates pattern matching},
  author={Shibata, Yusuxke and Kida, Takuya and Fukamachi, Shuichi and Takeda, Masayuki and Shinohara, Ayumi and Shinohara, Takeshi and Arikawa, Setsuo},
  year={1999},
  publisher={Technical Report DOI-TR-161, Department of Informatics, Kyushu University}
}

@article{song2020fast,
  title={Fast wordpiece tokenization},
  author={Song, Xinying and Salcianu, Alex and Song, Yang and Dopson, Dave and Zhou, Denny},
  journal={arXiv preprint arXiv:2012.15524},
  year={2020}
}

@misc{openai2022textembeddingada002,
  author       = {OpenAI},
  title        = {text-embedding-ada-002},
  year         = {2022},
  howpublished = {\url{https://platform.openai.com/docs/models/text-embedding-ada-002}},
  note         = {Accessed: 2025-07-18}
}

@article{raffel2020exploring,
  title={Exploring the limits of transfer learning with a unified text-to-text transformer},
  author={Raffel, Colin and Shazeer, Noam and Roberts, Adam and Lee, Katherine and Narang, Sharan and Matena, Michael and Zhou, Yanqi and Li, Wei and Liu, Peter J},
  journal={Journal of machine learning research},
  volume={21},
  number={140},
  pages={1--67},
  year={2020}
}

@article{douze2024faiss,
  title={The faiss library},
  author={Douze, Matthijs and Guzhva, Alexandr and Deng, Chengqi and Johnson, Jeff and Szilvasy, Gergely and Mazar{\'e}, Pierre-Emmanuel and Lomeli, Maria and Hosseini, Lucas and J{\'e}gou, Herv{\'e}},
  journal={arXiv preprint arXiv:2401.08281},
  year={2024}
}

@article{thakur2021beir,
  title={Beir: A heterogenous benchmark for zero-shot evaluation of information retrieval models},
  author={Thakur, Nandan and Reimers, Nils and R{\"u}ckl{\'e}, Andreas and Srivastava, Abhishek and Gurevych, Iryna},
  journal={arXiv preprint arXiv:2104.08663},
  year={2021}
}

@inproceedings{frobe2025corpus,
  title={Corpus Subsampling: Estimating the Effectiveness of Neural Retrieval Models on Large Corpora},
  author={Fr{\"o}be, Maik and Parry, Andrew and Scells, Harrisen and Wang, Shuai and Zhuang, Shengyao and Zuccon, Guido and Potthast, Martin and Hagen, Matthias},
  booktitle={European Conference on Information Retrieval},
  pages={453--471},
  year={2025},
  organization={Springer}
}

@article{wadden2020fact,
  title={Fact or fiction: Verifying scientific claims},
  author={Wadden, David and Lin, Shanchuan and Lo, Kyle and Wang, Lucy Lu and van Zuylen, Madeleine and Cohan, Arman and Hajishirzi, Hannaneh},
  journal={arXiv preprint arXiv:2004.14974},
  year={2020}
}

@article{kwiatkowski2019natural,
  title={Natural questions: a benchmark for question answering research},
  author={Kwiatkowski, Tom and Palomaki, Jennimaria and Redfield, Olivia and Collins, Michael and Parikh, Ankur and Alberti, Chris and Epstein, Danielle and Polosukhin, Illia and Devlin, Jacob and Lee, Kenton and others},
  journal={Transactions of the Association for Computational Linguistics},
  volume={7},
  pages={453--466},
  year={2019},
  publisher={MIT Press One Rogers Street, Cambridge, MA 02142-1209, USA journals-info~…}
}

@article{nguyen2016ms,
  title={Ms marco: A human-generated machine reading comprehension dataset},
  author={Nguyen, Tri and Rosenberg, Mir and Song, Xia and Gao, Jianfeng and Tiwary, Saurabh and Majumder, Rangan and Deng, Li},
  year={2016}
}

@inproceedings{boteva2016full,
  title={A full-text learning to rank dataset for medical information retrieval},
  author={Boteva, Vera and Gholipour, Demian and Sokolov, Artem and Riezler, Stefan},
  booktitle={European Conference on Information Retrieval},
  pages={716--722},
  year={2016},
  organization={Springer}
}

@misc{nutritionfacts2025,
  author       = {{NutritionFacts.org}},
  title        = {NutritionFacts.org},
  howpublished = {\url{https://nutritionfacts.org/}},
  year         = {2025},
  note         = {Accessed: 2025-07-19}
}

@article{yang2018hotpotqa,
  title={HotpotQA: A dataset for diverse, explainable multi-hop question answering},
  author={Yang, Zhilin and Qi, Peng and Zhang, Saizheng and Bengio, Yoshua and Cohen, William W and Salakhutdinov, Ruslan and Manning, Christopher D},
  journal={arXiv preprint arXiv:1809.09600},
  year={2018}
}

@inproceedings{maia201818,
  title={Www'18 open challenge: financial opinion mining and question answering},
  author={Maia, Macedo and Handschuh, Siegfried and Freitas, Andr{\'e} and Davis, Brian and McDermott, Ross and Zarrouk, Manel and Balahur, Alexandra},
  booktitle={Companion proceedings of the the web conference 2018},
  pages={1941--1942},
  year={2018}
}

@inproceedings{wachsmuth2018retrieval,
  title={Retrieval of the best counterargument without prior topic knowledge},
  author={Wachsmuth, Henning and Syed, Shahbaz and Stein, Benno},
  booktitle={Proceedings of the 56th Annual Meeting of the Association for Computational Linguistics (Volume 1: Long Papers)},
  pages={241--251},
  year={2018}
}

@misc{iyer2017quoraquestionpairs,
  author       = {Iyer, Shankar and Dandekar, Nikhil and Csernai, Kornél},
  title        = {First Quora Dataset Release: Question Pairs},
  howpublished = {Quora Data Blog},
  year         = {2017},
}

@article{thorne2018fever,
  title={FEVER: a large-scale dataset for fact extraction and VERification},
  author={Thorne, James and Vlachos, Andreas and Christodoulopoulos, Christos and Mittal, Arpit},
  journal={arXiv preprint arXiv:1803.05355},
  year={2018}
}

@article{diggelmann2020climate,
  title={Climate-fever: A dataset for verification of real-world climate claims},
  author={Diggelmann, Thomas and Boyd-Graber, Jordan and Bulian, Jannis and Ciaramita, Massimiliano and Leippold, Markus},
  journal={arXiv preprint arXiv:2012.00614},
  year={2020}
}

@misc{qwq32b,
    title = {QwQ-32B: Embracing the Power of Reinforcement Learning},
    url = {https://qwenlm.github.io/blog/qwq-32b/},
    author = {Qwen Team},
    month = {March},
    year = {2025}
}

@article{jarvelin2002cumulated,
  title={Cumulated gain-based evaluation of IR techniques},
  author={J{\"a}rvelin, Kalervo and Kek{\"a}l{\"a}inen, Jaana},
  journal={ACM Transactions on Information Systems (TOIS)},
  volume={20},
  number={4},
  pages={422--446},
  year={2002},
  publisher={ACM New York, NY, USA}
}

@inproceedings{voorhees1999trec,
  title={The TREC-8 Question Answering Track Evaluation.},
  author={Voorhees, Ellen M and Tice, Dawn M and others},
  booktitle={TREC},
  volume={1999},
  pages={82},
  year={1999}
}

@inproceedings{papineni2002bleu,
  title={Bleu: a method for automatic evaluation of machine translation},
  author={Papineni, Kishore and Roukos, Salim and Ward, Todd and Zhu, Wei-Jing},
  booktitle={Proceedings of the 40th annual meeting of the Association for Computational Linguistics},
  pages={311--318},
  year={2002}
}

@inproceedings{lin2004rouge,
  title={Rouge: A package for automatic evaluation of summaries},
  author={Lin, Chin-Yew},
  booktitle={Text summarization branches out},
  pages={74--81},
  year={2004}
}

@inproceedings{banerjee2005meteor,
  title={METEOR: An automatic metric for MT evaluation with improved correlation with human judgments},
  author={Banerjee, Satanjeev and Lavie, Alon},
  booktitle={Proceedings of the acl workshop on intrinsic and extrinsic evaluation measures for machine translation and/or summarization},
  pages={65--72},
  year={2005}
}

@article{han2023comprehensive,
  title={A comprehensive survey on vector database: Storage and retrieval technique, challenge},
  author={Han, Yikun and Liu, Chunjiang and Wang, Pengfei},
  journal={arXiv preprint arXiv:2310.11703},
  year={2023}
}

@article{li2023sentence,
  title={Sentence embedding leaks more information than you expect: Generative embedding inversion attack to recover the whole sentence},
  author={Li, Haoran and Xu, Mingshi and Song, Yangqiu},
  journal={arXiv preprint arXiv:2305.03010},
  year={2023}
}

@misc{pinecone,
  author       = {Pinecone},
  title        = {Pinecone},
  howpublished = {\url{https://www.pinecone.io}},
  note         = {Accessed: 2025-08-02},
  year         = 2025
}

@misc{milvus,
  author       = {Milvus},
  title        = {Milvus},
  howpublished = {\url{https://milvus.io}},
  note         = {Accessed: 2025‑08‑02},
  year         = 2025
}

@article{nogueira2019document,
  title={Document expansion by query prediction},
  author={Nogueira, Rodrigo and Yang, Wei and Lin, Jimmy and Cho, Kyunghyun},
  journal={arXiv preprint arXiv:1904.08375},
  year={2019}
}

@article{nogueira2019doc2query,
  title={From doc2query to docTTTTTquery},
  author={Nogueira, Rodrigo and Lin, Jimmy and Epistemic, AI},
  journal={Online preprint},
  volume={6},
  number={2},
  year={2019}
}

@article{wang2023query2doc,
  title={Query2doc: Query expansion with large language models},
  author={Wang, Liang and Yang, Nan and Wei, Furu},
  journal={arXiv preprint arXiv:2303.07678},
  year={2023}
}

@article{achiam2023gpt,
  title={Gpt-4 technical report},
  author={Achiam, Josh and Adler, Steven and Agarwal, Sandhini and Ahmad, Lama and Akkaya, Ilge and Aleman, Florencia Leoni and Almeida, Diogo and Altenschmidt, Janko and Altman, Sam and Anadkat, Shyamal and others},
  journal={arXiv preprint arXiv:2303.08774},
  year={2023}
}

@article{team2023gemini,
  title={Gemini: a family of highly capable multimodal models},
  author={Team, Gemini and Anil, Rohan and Borgeaud, Sebastian and Alayrac, Jean-Baptiste and Yu, Jiahui and Soricut, Radu and Schalkwyk, Johan and Dai, Andrew M and Hauth, Anja and Millican, Katie and others},
  journal={arXiv preprint arXiv:2312.11805},
  year={2023}
}

@article{team2024gemma,
  title={Gemma: Open models based on gemini research and technology},
  author={Team, Gemma and Mesnard, Thomas and Hardin, Cassidy and Dadashi, Robert and Bhupatiraju, Surya and Pathak, Shreya and Sifre, Laurent and Rivi{\`e}re, Morgane and Kale, Mihir Sanjay and Love, Juliette and others},
  journal={arXiv preprint arXiv:2403.08295},
  year={2024}
}

@misc{google-search,
  author       = {{Google}},
  title        = {Google Search},
  howpublished = {\url{https://www.google.com/}},
  note         = {Accessed: 2025-10-02}
}

@misc{bing-search,
  author       = {{Microsoft}},
  title        = {Bing Search},
  howpublished = {\url{https://www.bing.com/}},
  note         = {Accessed: 2025-10-02}
}

@misc{grammarly,
  author       = {{Grammarly Inc.}},
  title        = {Grammarly},
  howpublished = {\url{https://www.grammarly.com/}},
  note         = {Accessed: 2025-10-02}
}

@software{claude-code,
  author = {Anthropic},
  title = {Claude Code},
  year = {2026},
  url = {https://claude.ai},
}

@software{codex,
  author    = {OpenAI},
  title     = {Codex},
  year      = {2026},
  url       = {https://openai.com},
}

@software{hermes_agent,
  author       = {Nous Research},
  title        = {Hermes Agent},
  year         = {2026},
  url          = {https://hermes-agent.nousresearch.com/},
}

@software{open_claw,
  author = {OpenClaw},
  title = {OpenClaw },
  year = {2026},
  url = {https://openclaw.ai/},
}

@article{wang2025catch,
  title={Catch-only-one: Non-transferable examples for model-specific authorization},
  author={Wang, Zihan and Ma, Zhiyong and Ma, Zhongkui and Liu, Shuofeng and Liu, Akide and Wang, Derui and Xue, Minhui and Bai, Guangdong},
  journal={arXiv preprint arXiv:2510.10982},
  year={2025}
}

@inproceedings{wang2026re,
  title={Re-key-free, risky-free: Adaptable model usage control},
  author={Wang, Zihan and Ma, Zhongkui and Feng, Xinguo and Yan, Chuan and Liu, Dongge and Sun, Ruoxi and Wang, Derui and Xue, Minhui and Bai, Guangdong},
  booktitle={2026 IEEE 11th European Symposium on Security and Privacy (EuroS\&P)},
  pages={696--711},
  year={2026},
  organization={IEEE}
}

@article{ma2025convex,
  title={Convex hull approximation for activation functions},
  author={Ma, Zhongkui and Wang, Zihan and Bai, Guangdong},
  journal={Proceedings of the ACM on Programming Languages},
  volume={9},
  number={OOPSLA2},
  pages={1007--1033},
  year={2025},
  publisher={ACM New York, NY, USA}
}

@inproceedings{wang2025ai,
  title={Ai model modulation with logits redistribution},
  author={Wang, Zihan and Ma, Zhongkui and Feng, Xinguo and Mei, Zhiyang and Ma, Ethan and Wang, Derui and Xue, Minhui and Bai, Guangdong},
  booktitle={Proceedings of the ACM on Web Conference 2025},
  pages={4699--4709},
  year={2025}
}

@inproceedings{wang2024corelocker,
  title={Corelocker: Neuron-level usage control},
  author={Wang, Zihan and Ma, Zhongkui and Feng, Xinguo and Sun, Ruoxi and Wang, Hu and Xue, Minhui and Bai, Guangdong},
  booktitle={2024 IEEE symposium on security and privacy (SP)},
  pages={2497--2514},
  year={2024},
  organization={IEEE}
}
